\documentclass[runningheads]{llncs}
\usepackage[T1]{fontenc}
\usepackage{graphicx}
\usepackage{color}

\usepackage{cite}
\usepackage{amsmath}
\usepackage{amsfonts}
\usepackage{amssymb}
\usepackage{lscape}
\usepackage{tikz}
\usepackage{parskip}
\usepackage[shortlabels]{enumitem}
\usepackage{algorithm}
\usepackage{algpseudocode}
\usepackage{textcomp}
\usepackage{multirow}
\usepackage{caption}
\usepackage{subcaption}
\usepackage{xcolor}
\usepackage{ragged2e}

\begin{document}
\title{Probabilistic AutoRegressive Neural Networks for Accurate Long-range Forecasting}
\titlerunning{PARNN for long-range forecasting}
% If the paper title is too long for the running head, you can set
% an abbreviated paper title here
%
\author{Madhurima Panja\inst{1} \and
Tanujit Chakraborty\inst{1,2} \and
Uttam Kumar\inst{1}\and
Abdenour Hadid\inst{3}}
\authorrunning{Panja et al.}

\institute{Center for Data Sciences, International Institute of Information Technology Bangalore, India\\
\email{\{madhurima.panja, uttam\}@iiitb.ac.in} \\ \and 
Department of Science and Engineering, Sorbonne University Abu Dhabi, UAE\\
\email{tanujit.chakraborty@sorbonne.ae} \\ \and 
Sorbonne Center for Artificial Intelligence, Sorbonne University Abu Dhabi, UAE\\
\email{abdenour.hadid@sorbonne.ae}}
\maketitle              
\begin{abstract}
Forecasting time series data is a critical area of research with applications spanning from stock prices to early epidemic prediction. While numerous statistical and machine learning methods have been proposed, real-life prediction problems often require hybrid solutions that bridge classical forecasting approaches and modern neural network models. In this study, we introduce the Probabilistic AutoRegressive Neural Networks (PARNN), capable of handling complex time series data exhibiting non-stationarity, nonlinearity, non-seasonality, long-range dependence, and chaotic patterns. PARNN is constructed by improving autoregressive neural networks (ARNN) using autoregressive integrated moving average (ARIMA) feedback error, combining the explainability, scalability, and "white-box-like" prediction behavior of both models. Notably, the PARNN model provides uncertainty quantification through prediction intervals, setting it apart from advanced deep learning tools. Through comprehensive computational experiments, we evaluate the performance of PARNN against standard statistical, machine learning, and deep learning models, including Transformers, NBeats, and DeepAR. Diverse real-world datasets from macroeconomics, tourism, epidemiology, and other domains are employed for short-term, medium-term, and long-term forecasting evaluations. Our results demonstrate the superiority of PARNN across various forecast horizons, surpassing the state-of-the-art forecasters. The proposed PARNN model offers a valuable hybrid solution for accurate long-range forecasting. By effectively capturing the complexities present in time series data, it outperforms existing methods in terms of accuracy and reliability. The ability to quantify uncertainty through prediction intervals further enhances the model's usefulness in decision-making processes.

\keywords{Time Series Forecasting, ARIMA, Autoregressive neural networks, Hybrid model.}
\end{abstract}
\section{Introduction}
Time series forecasting has been a potential arena of research for the last several decades. Predicting the future is fundamental in various applied domains like economics, healthcare, demography, and energy, amongst many others, as it aids in better decision-making and formulating data-driven business strategies \cite{hyndman2007robust, chakraborty2020unemployment, panja2023epicasting}. Forecasts for any time series arising from different domains are usually obtained by modeling their corresponding data-generating process based on the analysis of the historical data. With the abundance of past data, there has been an increasing demand for forecasts in the industrial sectors as well \cite{xu2019forecasting,karmy2019hierarchical}. To serve the need to generate accurate and reliable forecasts, numerous statistical and machine learning methods have been proposed in the literature \cite{hyndman2008forecasting, hyndman2018forecasting, chakraborty2019forecasting, wang2022forecast}. One of the most popular statistical forecasting models is the autoregressive integrated moving average (ARIMA) \cite{box1970distribution} model that tracks the linearity of a stationary data-generating process. The ARIMA model describes the historical patterns in a time series as a linear combination of autoregressive (lagged inputs) and moving average (lagged errors) components. Despite its vast applicability, ARIMA is not suitable for modeling nonlinear datasets, as the model assumes the future values of the series to be linearly dependent on the past and current values. Thus, the modeling capability of linear models like ARIMA shrinks while dealing with real-world datasets that exhibit complex nonlinear and chaotic patterns \cite{zhang2003time}.

Modern tools of deep learning such as multilayer perceptron (MLP) \cite{rumelhart1986learning}, auto-regressive neural network (ARNN) \cite{faraway1998time}, and ensemble deep learning \cite{ray2021optimized} methods leverage the ground truth data (with nonlinear trends) to learn the temporal patterns in an automated manner. However, these methods fail to model long-term dependencies in time series. Current progress in computationally intelligent frameworks has brought us deep autoregressive (DeepAR) \cite{salinas2020deepar} model, neural basis expansion analysis (NBeats) \cite{oreshkin2019n}, temporal convolutional network (TCN) \cite{chen2020probabilistic}, Transformers \cite{wu2020deep}, and many others for modeling nonlinear and non-stationary datasets \cite{egrioglu2015recurrent, selvin2017stock}. The innovation in these advanced deep learning frameworks has demonstrated tremendous success in modeling and extrapolating the long-range dependence and interactions in temporal data; hence they are pertinent in the current literature on time series forecasting. Albeit these models are applied in several forecasting applications, their accuracy depends largely on the appropriate choice of hyperparameters, which are commonly data-dependent. Any misspecification of these parameters may lead to a significant reduction of the forecast accuracy for out-of-sample predictions. Moreover, these complex models often suffer from the problem of overfitting the given time series data, i.e., it learns the pattern and the noise in the underlying data to such an extent that it negatively impacts the prediction capability in the unseen data \cite{geman1992neural}. Another major drawback of some of the approaches mentioned above is the lack of explainability and the ``black-box-like'' behavior of these models.

To overcome the limitations arising from the stand-alone forecasting methods and to simplify the model selection procedure, several hybrid forecasting techniques that decompose a given time series into linear and nonlinear components have been proposed in the literature \cite{zhang2003time,babu2014moving,chakraborty2019forecasting}. The hybrid architectures comprise three steps: firstly, the linear patterns of the series are forecasted using linear models which are followed by an error re-modeling step using nonlinear models, and finally, the forecasts from both the steps are combined to produce the final output. These hybrid systems have been extended for forecasting applications in various domains such as finance \cite{cao2019financial,chakraborty2020unemployment}, earth science \cite{vautard2001validation,grover2015deep}, transportation \cite{xu2019forecasting}, energy \cite{qin2019hybrid,dong2016research}, agriculture \cite{shahwan2007forecasting}, epidemiology \cite{bhattacharyya2022stochastic,chakraborty2020real} and their forecasting performance has surpassed their counterpart models. Although the hybrid forecasting models have outperformed their counterpart models along with other individual baseline forecasters in various applications, they are constructed based on certain assumptions. For instance, it is assumed that a series's linear and nonlinear patterns can be modeled separately or that the residuals comprise only the nonlinear trends. Alternatively, an additive or multiplicative relationship exists between the linear and nonlinear segments of the datasets \cite{chakraborty2020real}. However, if these assumptions do not hold, i.e., if the existing linear and nonlinear components cannot be modeled separately or if the residuals of the linear model do not contain valid nonlinear patterns, or if there is a lack of linear relationship between the components of a series, then the forecast accuracy of these hybrid models might substantially degrade. Since most real-world time series datasets may not satisfy any of these assumptions, the use of these hybrid frameworks may not be adequate in such scenarios. 

Motivated by the above observations, we propose a hybrid probabilistic autoregressive neural networks (PARNN) that is designed to overcome the limitations of hybrid time series forecasting models while improving their predictive accuracies. The first phase of our proposal encompasses a linear ARIMA model fitting and generating in-sample residuals. During the second phase, these unexplained residuals along with the original input lagged series are remodeled using a nonlinear ARNN model. In the proposed hybridization, neural network autoregression is used to model not only the time series data but also the feedback errors of the linear ARIMA model to improve the learning ability of the network. This combined architecture is easier to train and can accurately make long-range predictions for a wide variety of irregularities of real-world complex time series. The proposed framework is extensively evaluated on twelve publicly available benchmark time series datasets. The experimental results show that our proposed model provides accurate and robust predictions for nonlinear, non-stationary, chaotic, and non-Gaussian data, outperforming several statistical, machine learning, deep learning, hybrid, and ensemble forecasters in most real-data problems.

\section{Background}
Most time series models tend to suffer in modeling the complexities of real-world datasets. The performance of hybrid models in such situations is encouraging. We aim to improve the forecasting performance of hybrid models by proposing a hybrid PARNN model. Our proposed model comprises the linear ARIMA model and the nonlinear ARNN model. In this section, we briefly describe the constituent models along with popularly used hybrid and ensemble models before describing our proposed framework.
%This work proposes hybrid PARNN model for time series forecasting. The proposal uses statistical ARIMA model and machine-learning based ARNN model in its framework. Before describing the proposed methods, we briefly describe the individual models to be used in the hybridization.

\subsection{ARIMA Model}
The classical autoregressive integrated moving average (ARIMA) model, often termed as Box-Jenkins method \cite{box1970distribution}, is one of the most widely used statistical models in the forecasting literature. The ARIMA$(p,d,q)$ model comprises three parameters where $p$ and $q$ denote the order of the AR and MA terms, and $d$ denotes the order of differencing. The mathematical formulation of the ARIMA model is given by
\begin{equation*}
  y_t = \beta_0 +\alpha_1 y_{t-1}+\alpha_2 y_{t-2}+\ldots+\alpha_p y_{t-p} +\varepsilon_{t}-\beta_1\varepsilon_{t-1}-\beta_2\varepsilon_{t-2}-\ldots-\beta_q\varepsilon_{t-q}
\end{equation*}
where, $y_t$ is the actual time series, $\varepsilon_t$ is the random error at time $t$, and $\alpha_i$ and $\beta_j$ are the model parameters.
The ARIMA model is constructed using three iterative steps. Firstly, we convert a non-stationary series into a stationary one by applying the difference of order $d$. Then, once a stationary series is obtained, we select the model parameters $p$ and $q$ from the ACF plot and PACF plot, respectively. Finally, we obtain the ``best fitted'' model by analyzing the residuals. 

\subsection{ARNN Model}
Autoregressive neural networks (ARNN), derived from the artificial neural network (ANN), is specifically designed for modeling nonlinear time series data \cite{faraway1998time}. The ARNN model comprises a single hidden layer embedded within its input and output layers. The ARNN$(u,v)$ model passes $u$ lagged input values from its input layer to the hidden layer comprising of $v$ hidden neurons. The value of $v$ is determined using the formula $v=\lceil(u+1)/2)\rceil$, where $\lceil\cdot\rceil$ is the ceiling function (also known as least integer function) as proposed in \cite{hyndman2018forecasting, panja2023epicasting}. After being trained by a gradient descent back-propagation approach \cite{rumelhart1986learning}, the final forecast is obtained as a linear combination of its predictors. The mathematical formulation of the ARNN model is given by:
\begin{equation*}
	g(\underline{y}) = \alpha_{0}^* + \displaystyle \sum_{j=1}^{v} \alpha_j^* \phi \left( \beta_j^* + \theta_{j}^{*'}\underline{y} \right),
\end{equation*}
where $g$ denotes a neural network, $\underline{y}$ is a $u$-lagged inputs, $\alpha_0^*, \beta_j^*, \alpha_j^*$ are connecting weights, $\theta_j^*$ is a weight vector of dimension $u$ and $\phi$ is a bounded nonlinear activation function. %For practical implementation, the 'nnetar'  function available in the ``forecast" package in R statistical software implements ARNN model \cite{hyndman2020package}.

\subsection{Ensemble and hybrid models}
In the context of forecasting literature, although ARIMA and ARNN models have individually achieved significant successes in their respective domains \cite{kodogiannis2002forecasting,nochai2006arima}, their difficulties in modeling the complex autocorrelation structures within real-world datasets have led to the development of ensemble and hybrid forecasting approaches. The idea of the ensemble forecasting model was first introduced in 1969 \cite{bates1969combination}. The final output of these models was a weighted combination of the forecasts generated by its component models i.e., $$\hat{y} = \sum_{i \in \mathcal{M}}{\gamma_{i}\hat{y}_i},$$ where $\gamma_i$ denotes the weights and $\hat{y}_i$ denotes the forecast generated from the $i^{th}, \; i = 1, 2, \ldots, \mathcal{M}$ individual model. The ensemble models could significantly improve forecast accuracy for various applications. However, the appropriate selection of weights, often termed as ``forecast combination puzzle'' posed a significant challenge to its universal success. 

The concept of hybrid forecasting was led by Zhang in 2003 \cite{zhang2003time} where the author assumed that a time series could be decomposed into its linear and nonlinear components as follows:
$$y_t = y_{Lin_t} + y_{Nlin_t},$$ where $y_{Lin_t}$ and $y_{Nlin_t}$ denote the linear and nonlinear elements, respectively. Using a linear model (e.g., ARIMA), $y_{Lin_t}$ was estimated from the available dataset. Let the forecast value of the linear model be denoted by $\hat{y}_{Lin_t}$. Then, the model residuals, $y_{Res_t}$ were computed as: $$y_{Res_t} = y_t - \hat{y}_{Lin_t}.$$ The left-out autocorrelations in the residuals were further re-modelled using a nonlinear model (e.g., ANN or ARNN) which generates the nonlinear forecasts as $\hat{y}_{Nlin_t}$. The final forecast $\hat{y}_t$ is computed by adding the out-of-sample forecasts generated by the linear and the nonlinear models i.e., $$\hat{y}_t = \hat{y}_{Lin_t} + \hat{y}_{Nlin_t}.$$ 
Various combinations of hybrid models have been proposed in the literature, showing forecasting performance improvement \cite{panigrahi2017hybrid, bhattacharyya2021theta, dave2021forecasting}. Among these models, ARIMA-ARNN models, for instance, have shown higher accuracies for epidemic \cite{chakraborty2019forecasting} and econometric modeling \cite{chakraborty2020unemployment}. %Furthermore, practical implementation of several ensemble models can be done using an R package called ``forecastHybrid" \cite{shaub2019forecasthybrid} and for hybrid models, refer to \cite{chakraborty2020nowcasting}.

\section{Proposed Model}\label{PARNN_Model_Section}
In this section, we discuss the formulation of the proposed PARNN model and the computation of its prediction intervals. Practical usage of the proposal for real-world time series datasets is presented in Sec. \ref{experimental}.

\subsection{PARNN model}
\begin{figure*}
	\centering
	\includegraphics[scale=0.3]{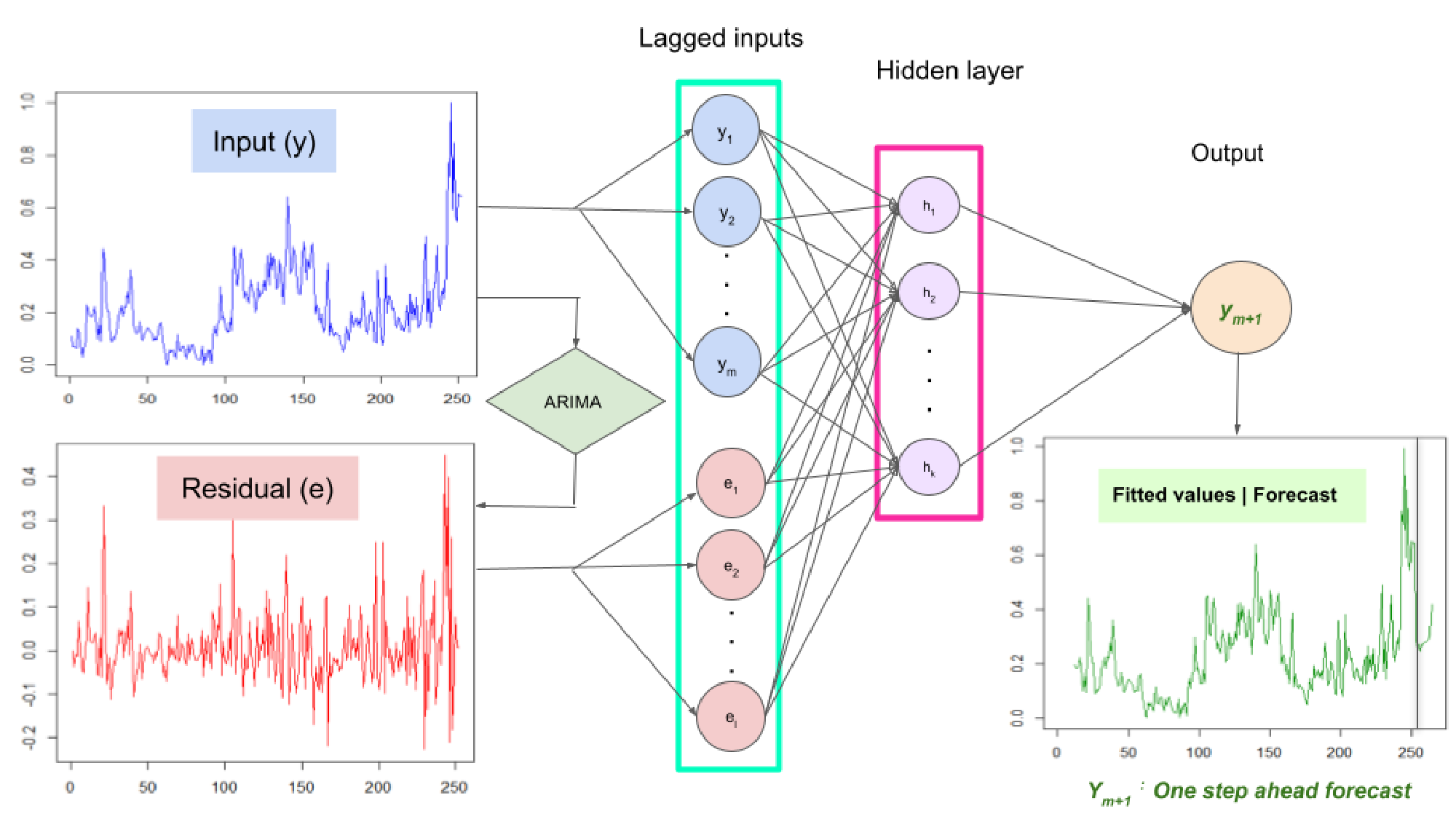}	\caption{Schematic diagram of proposed PARNN model, where $y_t$ and $e_t$ denote the original time series and residual series obtained using ARIMA, respectively. The value of $k$ in the hidden layer is set to $k=\lceil(m+l+1)/2\rceil$.}
	\label{model}
\end{figure*}

We propose a hybrid probabilistic autoregressive neural networks (PARNN) which is a modification to the artificial neural network (ANN$(p,d,q)$) \cite{khashei2010artificial} and recurrent neural network (RNN) \cite{connor1994recurrent} models used for nonlinear time series forecasting. The current hybrid models in time series \cite{zhang2003time}, such as the hybrid ARIMA-ARNN model \cite{chakraborty2019forecasting}, made a significant improvement in the predictive accuracy of individual forecasters like ARIMA and ARNN models. These hybrid models assume an additive relationship between linear and nonlinear components of the time series which is not always obvious. This paper builds a probabilistic ARNN framework that blends classical time series with scalable neural network architecture. Our proposed PARNN$(m,k,l)$ model considers the future values of the time series to be a nonlinear function of $m$-lagged values of past observations and $l$-lagged values of ARIMA residuals (feedback errors). Therefore, we have, 
$$ y_t = f(y_{t-1},\ldots,y_{t-m}, e_{t-1},\ldots,e_{t-l}),$$
where the nonlinear function $f$ is a single hidden-layered autoregressive neural networks having $k$ hidden neurons, $y_t$ and $e_t$ denote the actual time series and the residual series at time $t$. Using the connection weights $\alpha$ and a bounded nonlinear sigmoidal activation function $G(\cdot)$ of the neural network, $y_t$ can be denoted as
\begin{equation}
y_t = \alpha_{0} +\displaystyle \sum_{j=1}^{k} \alpha_j G \left(\alpha_{0,j} + \sum_{i=1}^{m} \alpha_{i,j}y_{t-i} + \sum_{i=m+1}^{m+l} \alpha_{i,j} e_{t+m-i}\right) + \epsilon_t.
\label{eq1}
\end{equation}
The learning phase of the proposed PARNN model comprises two stages. In the first stage, the original series is standardized and the ARIMA residuals $(e_t)$ are generated from the standardized series $(y_t)$ using the best-fitted ARIMA model with minimum Akaike Information Criterion (AIC). In the second stage, a single hidden layered feed-forward neural network is built to model the nonlinear and linear relationships using ARIMA residuals and original observations. This neural network receives $m$ lagged values of $y_t$ and $l$  lagged values of $e_t$ as inputs. We denote the proposed model as PARNN$(m,k,l)$, where $k$ is the number of neurons arranged in a single hidden layer and set to $k = \lceil\frac{m+l+1}{2}\rceil$ (where $\lceil \cdot \rceil$ is the ceiling function) as in \cite{hyndman2018forecasting, panja2023epicasting}. The proposed network architecture generates a one-step-ahead forecast for the series $(y_t)$ at each iteration. The model employs historical observations and recent forecasts as inputs for generating multi-step ahead forecasts iteratively. The proposed PARNN is useful for modeling the underlying data-generating process with non-stationary, nonlinear, and non-Gaussian structures in the time series data. Nonlinear ARMA and simple ARNN models can be seen as special cases of our proposed model. It can be seen that in the proposed PARNN($m,k,l$) model, in contrast to the traditional hybrid models, no prior assumption is considered for the relationship between the linear and nonlinear components. Additionally, it can be generally guaranteed that the performance of the probabilistic ARNN model will not be worse than either of the components -- ARIMA and ARNN -- used separately or hybridized. We determine the optimal values of the hyperparameters $m$ and $l$ by performing a grid search algorithm on the validation dataset. The maximum values of both the hyperparameters are usually set to $10$, and the model is iterated over a grid of $m$ and $l$. At each iteration, the model is fitted for different pairs of values of $m$ and $l$, and the forecasts $\hat{y}_t$ are generated for the validation set (a part of the training set is kept separately for validation to avoid data leakage). The performance of these fitted models is evaluated for all possible pairs of $m$ and $l$ on the validation set, and eventually, the pair of values exhibiting the best performance w.r.t. the mean absolute scaled error (MASE) metric is selected. Thus, our proposed PARNN can be considered as a hybrid framework in which ARIMA is embedded into an ARNN model which converges very fast unlike the ANN$(p,d,q)$ and RNN models for nonlinear forecasting. The proposed PARNN model is preferable for long-range forecasting which is one of the most attractive features of autoregressive neural networks \cite{leoni2009long}. A workflow diagram of the proposed PARNN model is presented in Fig. \ref{model}. An algorithmic version of our proposal is given in Algorithm \ref{alg:cap}. 

\begin{algorithm}[!htp]
\caption{PARNN ($m,k,l$) model}\label{alg:cap}
\begin{algorithmic}
\item \textbf{Input:} The standardized time series $y_t$ 
\item \textbf{Output:} Forecasts of $y_t$ ($\hat{y}_{t+1}, \hat{y}_{t+2}, \ldots$) for a desired test horizon.
\item \textbf{Steps:}
\begin{enumerate}
    \item Split the series into in-sample (training and validation) and test data (out-of-sample).
    \item Fit an ARIMA($p,d,q$) model using the training data.
    \begin{itemize}
        \item Obtain the predicted values by using the ARIMA model.
        \item Generate the residuals $e_t$ by subtracting the ARIMA predicted values from the observed time series $y_t$.
    \end{itemize}
    \item  Fit a single hidden layered feed-forward neural network using the actual time series $y_t$ and ARIMA feedback residuals $e_t$. 
    \begin{itemize}
        \item Initiate the network with a random starting weight and provide $m$ lagged values of the original series and $l$ lagged values of the error as the input to the network with hidden neurons $k=\lceil(m+l+1)/2\rceil$ (where $\lceil \cdot \rceil$ is the ceiling function).
        \item We determine the optimal values of $m$ and $l$ by performing a grid search algorithm on the validation set (twice the size of the test set). Several PARNN models with performance measures are recorded. Eventually, the model with the least MASE score is selected.
        %\item The number of hidden nodes is set to $k=[(p+q+1)/2]$
    \end{itemize}
    \item Final forecast is acquired by considering the average output of 500 neural networks with optimized hyperparameters selected in the previous step and applying inverse standardization. 
    \item The above procedure generates a one-step-ahead forecast for the given series $y_t$. To generate the multi-step ahead forecasts we iteratively repeat the above procedure with the latest forecast as lagged input.
    
    \end{enumerate}
\end{algorithmic}
\end{algorithm}

\subsection{Prediction Interval of the PARNN model}
Owing to the hybrid nature of the proposed PARNN model, computation of the prediction interval cannot be done analytically, so we simulate the future sample path to interpret the uncertainty in the forecasts. The forecast of our proposed model at time $t$ can be denoted as
$$y_t = f(\mathbf{y_{t-1}}, \mathbf{e_{t-1}}) + \epsilon_t,$$ 
where, $\mathbf{y_{t-1}}= (y_{t-1},y_{t-2},\ldots,y_{t-m})'$ are the lagged inputs and $\mathbf{e_{t-1}}= (e_{t-1},e_{t-2},\ldots,e_{t-l})'$ are the lagged ARIMA feedback, and $f$ is the neural network with $k$ hidden nodes. We assume that the error, $\epsilon_t$ follows a Gaussian distribution and ensure that it is homoscedastic in nature (by applying Box-Cox transformation with parameter 0.5). We can then iteratively use random samples drawn from this distribution to simulate future sample paths of the model. Thus, if $\epsilon_{t+1}$ is the sample drawn at time $t+1$ then,
$$y_{t+1} = f(\mathbf{y_t}, \mathbf{e_t}) + \epsilon_{t+1},$$ 
where, $\mathbf{y_t}= (y_t,y_{t-1},\ldots,y_{t-m+1})'$ and $\mathbf{e_t}= (e_t,e_{t-1},\ldots,e_{t-l+1})'$. 
We repeat this simulation 1000 times at each time step in the forecast horizon and build up the knowledge about the distribution of the forecasted values. The 80\% prediction intervals are generated by calculating the corresponding percentiles from the distributions of the future values for the PARNN model, since for the higher prediction levels the intervals are wider and irrelevant to practical use cases.

\section{Experimental Analysis}\label{experimental}
In this section, we analyze the performance of our proposed PARNN model by conducting extensive experiments on twelve open-access time series datasets with varying frequencies and compare the results with other state-of-the-art forecasters from statistical, machine learning, and deep learning paradigms.

\subsection{Data}
The datasets used for the experimental evaluation exhibit distinct global characteristics and are collected from various applied domains as listed below:

\textbf{Finance and Economics:} In our study we consider the daily adjusted close price of three NASDAQ stock exchange datasets, namely Alphabet Inc Class C (\emph{GOOG Stock}), Microsoft Corporation (\emph{MSFT Stock}), and Amazon Inc. (\emph{AMZN Stock}) collected during the financial years 2020-2022 from Yahoo finance. Additionally, we use three macroeconomic datasets comprising of the economic policy uncertainty index for the United States during 2000-2021 (\emph{US-EPU Index}), the unemployment rate in the United Kingdom during the period 1971-2021 (\emph{UK Unemployment}), and the national exchange rate of Russian currency units per U.S. dollars (\emph{Russia Exchange}) during 2000-2021. These datasets provide monthly averages for the relevant series and are collected from the Fred Economic Data repository. Analysing the forecasting performance of the proposed framework for these financial and economic datasets is crucial because accurate forecasts in these domains aid in decision-making, enable risk management, guide policy formulation, promote market efficiency, and help understand future trends and adapt accordingly.

\textbf{Epidemics:} Alongside, economic datasets, we utilize four epidemiological data to assess the efficacy of our proposed PARNN model for predicting future infectious disease dynamics and guide health practitioners in devising appropriate interventions that mitigate the impacts of epidemics. These open-access datasets, collected from the CDC data repository, indicate the weekly crude incidence rate of dengue and malaria diseases in Colombia and Venezuela regions during 2005-2016 and 2002-2014, respectively. 

\textbf{Demography and Business:} Furthermore, we consider two benchmark data, namely \emph{Births} and \emph{Tourism} in our experimentation. The Births dataset, collected from the ``mosaicData'' package of R software, denotes the daily number of births occurring in the US during 1968-1988. On the contrary, the Tourism dataset indicates the number of domestic holiday trips happening in Melbourne, Australia reported quarterly from 1998 to 2016, publicly available at \cite{hyndman2018forecasting}. Generating accurate forecasts for these datasets is essential for practitioners to adapt, seize opportunities, and effectively address challenges in their respective domains. Thus, the forecast efficiency of the proposed PARNN model in comparison with benchmarks in various applied domains is crucial in our study.

\subsection{Analysing Global Characteristics in the Datasets}
Twelve open-access time series datasets considered in this study are free from missing observations and are of different lengths. We have examined several global characteristics such as long-term dependency (LTD), non-stationarity, nonlinearity, seasonality, trend, normality, and chaotic behavior of these datasets using Hurst exponent, KPSS test, Ter{\"a}svirta test (nonlinearity test), Ollech and Webel's combined seasonality test, Mann-Kendall test, Anderson-Darling test, and Lyapunov exponent, respectively, to understand their structural patterns \cite{hyndman2018forecasting, chakraborty2020nowcasting}. We performed all the statistical tests using R statistical software. The global characteristics of these datasets evaluated based on the above-stated statistical measures are summarized in Table \ref{Data_description}.

\begin{table*}[!ht]
\caption{Global characteristics of the real-world datasets (green circle indicates the presence of feature and the red circle indicates the absence of feature).}
  \centering \scriptsize
  \begin{tabular}{|c|c|c|c|c|c|c|c|c|c|c|}
         \hline
        Datasets &	Frequency &	Time Span &	Length & LTD & Stationary & Linear & Seasonal & Trend & Gaussian & Chaotic\\ \hline
        GOOG Stock & \multirow{4}{*}{Daily} & 2020-2022 & 504 & \tikz\draw[green,fill=green] (0,0) circle (.7ex); & \tikz\draw[red,fill=red] (0,0) circle (.7ex); & \tikz\draw[red,fill=red] (0,0) circle (.7ex); & \tikz\draw[red,fill=red] (0,0) circle (.7ex); &  \tikz\draw[green,fill=green] (0,0) circle (.7ex);  & \tikz\draw[red,fill=red] (0,0) circle (.7ex); & \tikz\draw[green,fill=green] (0,0) circle (.7ex); \\
        MSFT Stock & & 2020-2022 & 504 & \tikz\draw[green,fill=green] (0,0) circle (.7ex); & \tikz\draw[red,fill=red] (0,0) circle (.7ex); & \tikz\draw[green,fill=green] (0,0) circle (.7ex); & \tikz\draw[red,fill=red] (0,0) circle (.7ex); & \tikz\draw[green,fill=green] (0,0) circle (.7ex); & \tikz\draw[red,fill=red] (0,0) circle (.7ex); &  \tikz\draw[red,fill=red] (0,0) circle (.7ex); \\
        
        AMZN Stock  & & 2020-2022 & 504 & \tikz\draw[green,fill=green] (0,0) circle (.7ex); & 
        \tikz\draw[red,fill=red] (0,0) circle (.7ex); & \tikz\draw[green,fill=green] (0,0) circle (.7ex); & 
        \tikz\draw[red,fill=red] (0,0) circle (.7ex); & \tikz\draw[green,fill=green] (0,0) circle (.7ex); & 
        \tikz\draw[red,fill=red] (0,0) circle (.7ex); & \tikz\draw[green,fill=green] (0,0) circle (.7ex); \\
        
        Births & & 1968-1988 & 7305 & \tikz\draw[green,fill=green] (0,0) circle (.7ex); & 
        \tikz\draw[red,fill=red] (0,0) circle (.7ex); & \tikz\draw[red,fill=red] (0,0) circle (.7ex); & \tikz\draw[green,fill=green] (0,0) circle (.7ex);  & \tikz\draw[green,fill=green] (0,0) circle (.7ex); & \tikz\draw[red,fill=red] (0,0) circle (.7ex); & \tikz\draw[red,fill=red] (0,0) circle (.7ex); \\  \hline
        Colombia Dengue & \multirow{4}{*}{Weekly} & 2005-2016 & 626 & \tikz\draw[green,fill=green] (0,0) circle (.7ex); & \tikz\draw[red,fill=red] (0,0) circle (.7ex); &  \tikz\draw[red,fill=red] (0,0) circle (.7ex); &   \tikz\draw[green,fill=green] (0,0) circle (.7ex); &  \tikz\draw[green,fill=green] (0,0) circle (.7ex); & \tikz\draw[red,fill=red] (0,0) circle (.7ex); & \tikz\draw[green,fill=green] (0,0) circle (.7ex); \\
        Colombia Malaria && 2005-2016 & 626 & \tikz\draw[green,fill=green] (0,0) circle (.7ex); & \tikz\draw[red,fill=red] (0,0) circle (.7ex); &  \tikz\draw[green,fill=green] (0,0) circle (.7ex); &   \tikz\draw[green,fill=green] (0,0) circle (.7ex); &  \tikz\draw[green,fill=green] (0,0) circle (.7ex); & \tikz\draw[red,fill=red] (0,0) circle (.7ex); & \tikz\draw[green,fill=green] (0,0) circle (.7ex); \\
        Venezuela Dengue && 2002-2014 & 660 & \tikz\draw[green,fill=green] (0,0) circle (.7ex); & \tikz\draw[red,fill=red] (0,0) circle (.7ex); &  \tikz\draw[green,fill=green] (0,0) circle (.7ex); &   \tikz\draw[green,fill=green] (0,0) circle (.7ex); &  \tikz\draw[green,fill=green] (0,0) circle (.7ex); & \tikz\draw[red,fill=red] (0,0) circle (.7ex); & \tikz\draw[green,fill=green] (0,0) circle (.7ex); \\
        Venezuela Malaria && 2002-2014 & 669 & \tikz\draw[green,fill=green] (0,0) circle (.7ex); & \tikz\draw[red,fill=red] (0,0) circle (.7ex); &  \tikz\draw[red,fill=red] (0,0) circle (.7ex); &   \tikz\draw[green,fill=green] (0,0) circle (.7ex); &  \tikz\draw[green,fill=green] (0,0) circle (.7ex); & \tikz\draw[red,fill=red] (0,0) circle (.7ex); & \tikz\draw[green,fill=green] (0,0) circle (.7ex); \\ \hline
        US EPU Index & \multirow{3}{*}{Monthly} & 2000-2021 & 264 & \tikz\draw[green,fill=green] (0,0) circle (.7ex); & \tikz\draw[red,fill=red] (0,0) circle (.7ex); &  \tikz\draw[green,fill=green] (0,0) circle (.7ex); &   \tikz\draw[red,fill=red] (0,0) circle (.7ex); &  \tikz\draw[green,fill=green] (0,0) circle (.7ex); & \tikz\draw[red,fill=red] (0,0) circle (.7ex); & \tikz\draw[green,fill=green] (0,0) circle (.7ex); \\        
        UK unemployment && 1971-2016 & 552 & \tikz\draw[green,fill=green] (0,0) circle (.7ex); & \tikz\draw[red,fill=red] (0,0) circle (.7ex); &  \tikz\draw[green,fill=green] (0,0) circle (.7ex); &   \tikz\draw[red,fill=red] (0,0) circle (.7ex); &  \tikz\draw[red,fill=red] (0,0) circle (.7ex); & \tikz\draw[red,fill=red] (0,0) circle (.7ex); & \tikz\draw[red,fill=red] (0,0) circle (.7ex); \\    
        Russia Exchange && 2000-2021 & 264 & \tikz\draw[green,fill=green] (0,0) circle (.7ex); & \tikz\draw[red,fill=red] (0,0) circle (.7ex); &  \tikz\draw[red,fill=red] (0,0) circle (.7ex); &   \tikz\draw[red,fill=red] (0,0) circle (.7ex); &  \tikz\draw[green,fill=green] (0,0) circle (.7ex); & \tikz\draw[red,fill=red] (0,0) circle (.7ex); & \tikz\draw[green,fill=green] (0,0) circle (.7ex); \\ \hline
        Tourism & Quarterly & 1998-2017 & 80 & \tikz\draw[green,fill=green] (0,0) circle (.7ex); & \tikz\draw[red,fill=red] (0,0) circle (.7ex); &  \tikz\draw[green,fill=green] (0,0) circle (.7ex); &   \tikz\draw[red,fill=red] (0,0) circle (.7ex); &  \tikz\draw[green,fill=green] (0,0) circle (.7ex); & \tikz\draw[red,fill=red] (0,0) circle (.7ex); & \tikz\draw[green,fill=green] (0,0) circle (.7ex); \\ \hline
    \end{tabular}
    \label{Data_description}
\end{table*}

\subsection{Performance Measures}
We adopt the Mean Absolute Scaled Error (MASE), Root Mean Square Error (RMSE), and Symmetric Mean Absolute Percent Error (SMAPE) as the key performance indicators in our study \cite{hyndman2018forecasting}. The mathematical formulae of these metrics are given below:
{
\begin{align*}
MASE &= \frac{\sum_{i=M+1}^{M+n} |y_i - \hat{y}_i|}{\frac{n}{M-S}  \sum_{i=S+1}^{M} |y_i - y_{i-s}|}; \; RMSE = \left(\frac{1}{n} \sum_{i=1}^n (y_i - \hat{y}_i)^2\right)^\frac{1}{2}; \\ \nonumber
SMAPE &= \frac{1}{n} \sum_{i=1}^n  \frac{|y_i -\hat{y}_i|}{(|y_i|+|\hat{y}_i|)/2}*100\%;  \nonumber
\end{align*}}
\noindent where $y_i$ is the actual value, $\hat{y}_i$ is the predicted output, $n$ is the forecast horizon, $M$ denotes the training data size, and $S$ is the seasonality of the dataset. The model with the least error measure is considered the `best' model. 

\subsection{Baselines}
In this study, we have compared the performance of our proposed PARNN framework with several forecasters from different paradigms. These benchmarks include statistical methods, namely random walk with drift (RWD) \cite{entorf1997random}, exponential smoothing (ETS) \cite{winters1960forecasting}, autoregressive integrated moving average (ARIMA) \cite{box1970distribution}, and TBATS (T: Trigonometric. B: Box-Cox transformation. A: ARIMA errors. T: Trend. S: Seasonal components) \cite{de2011forecasting}; machine learning models -- multilayer perceptron (MLP) \cite{rumelhart1986learning}, autoregressive neural networks (ARNN) \cite{faraway1998time}; deep learning frameworks namely NBeats \cite{oreshkin2019n}, DeepAR \cite{salinas2020deepar}, Temporal convolutional networks (TCN) \cite{chen2020probabilistic}, and Transformers \cite{wu2020deep}, and hybrid models specifically hybrid ARIMA-ANN model (named as Hybrid-1) \cite{zhang2003time}, hybrid ARIMA-ARNN model (named as Hybrid-2) \cite{chakraborty2019forecasting}, and hybrid ARIMA-LSTM model (named as Hybrid-3) \cite{dave2021forecasting}.

\subsection{Model Implementations}
In this section, we discuss the implementation of the proposed PARNN framework and other state-of-the-art models. To assess the scalability of our proposal, we conduct the experimentation with three different forecast horizons (test data) as short-term, medium-term, and long-term spanning over $({5\%,\;10\%,\;20\%})$ observations for daily datasets, $({13,\;26,\;52})$ weeks for weekly datasets, $({6,\;12, \; 24})$ months for monthly datasets, and $({4,\;8,\;12})$ quarters for quarterly dataset, respectively. We begin the experiments by partitioning each dataset into a train, validation (twice the size of the test set), and test segments in chronological order. For the proposed PARNN model, we initially standardize the training data and fit an ARIMA model using the `auto.arima' function of R statistical software. The best-fitted ARIMA model was selected using the Akaike information criterion (AIC) and employed to estimate the residual values. Further, we apply an error correction approach to model the original time series along with the left-out ARIMA residuals using our proposed PARNN$(m,k,l)$ model. We provide a combined input comprising of $m$-lagged values of the original time series and $l$-lagged values of the ARIMA residuals to the PARNN$(m,k,l)$ model and implement the proposed framework as described in Section \ref{PARNN_Model_Section}.
%The training phase of the proposed model was started with random initialization and weights of the neural network were trained using a gradient descent back-propagation approach \cite{rumelhart1986learning}. The optimal values of the hyperparameters $m$ and $l$ are selected using a grid-search algorithm over the validation set, and the number of hidden neurons in the hidden layer is set to $k=\lceil(m+l+1)/2\rceil$ (where $\lceil \cdot \rceil$ is the ceiling function) \cite{hyndman2018forecasting, panja2023epicasting}. 
The optimal values of the hyperparameters $(m,k,l)$ of the proposed model are selected by minimizing the MASE score and are summarized for all the datasets in Table \ref{PARNN_parameters}. We iterate the training of the PARNN model over 500 repetitions with the chosen $(m,k,l)$ values and obtain the final forecast as the average output of these networks. For instance, in the Colombia Malaria dataset for medium-term forecast, ARIMA$(1,1,2)$ model was fitted and $AIC = 8168.85$ was obtained. Further, the model residuals and the training data were remodeled using the PARNN model, where each of the parameters $m$ and $l$ were tuned over a grid of values between $(1,10)$. This grid search yields a minimum MASE score at PARNN$(1,2,2)$ model. Following this, the fitted PARNN$(1,2,2)$ model was utilized to generate the final forecast as the average output of 500 networks. The predicted values are then used to evaluate the model performance. In a similar manner, we fit the PARNN model with optimal hyperparameters to the other datasets and generate the final forecast for desired horizons. Additionally, we have implemented the other individual statistical and machine learning models like RWD, ETS, ARIMA, TBATS, and ARNN models using the ``forecast'' package in R statistical software \cite{hyndman2020package}. The MLP framework is fitted using the ``nnfor'' package in R \cite{kourentzes2017nnfor}. For the deep learning frameworks, we have utilized the ``darts'' library in Python \cite{herzen2022darts}, and the hybrid models are fitted using the implementation provided in \cite{chakraborty2020nowcasting}. Owing to the variance of machine learning and deep learning models we repeat the experiments ten times and report the mean error measures and their respective standard deviation.

\begin{table*}[!ht]
\caption{Estimated parameter values of the proposed PARNN($m,k,l$) model for forecasting the pre-defined short (ST), medium (MT), and long (LT) term forecast horizons of the chosen time series data sets.}
\scriptsize
\centering
 \begin{tabular}{|c|cccccccccccc|}
%{|p{0.02\textwidth}|p{0.04\textwidth}|p{0.04\textwidth}|p{0.04\textwidth}|p{0.04\textwidth}|p{0.053\textwidth}|p{0.053\textwidth}|p{0.053\textwidth}|p{0.052\textwidth}|p{0.053\textwidth}|p{0.053\textwidth}|p{0.048\textwidth}|p{0.05\textwidth}|p{0.045\textwidth}|p{0.04\textwidth}|} 
\hline
Data  & GOOG & MSFT & AMZN & Births & Colombia & Colombia & Venezuela & Venezuela & US-EPU & UK & Russia & Tourism \\
& Stock & Stock & Stock && Dengue & Malaria & Dengue & Malaria & Index & Unemp. & Exchange & \\ \hline
ST & $2,2,1$ & $10,6,1$ & $1, 2,1$ & $6,6,4$ & 
$2, 2, 1$ & $3,3,1$ & $5,6,6$ & $3,6,7$ & 
$4,3,1$ & $10,7,3$ &  $3,3,1$ 
& $5,4,1$ \\ &&&&&&&&&&&& \\ \hline
MT &  $10, 8, 5$ & $7,7,5$ & $5,4,1$  & $7,5,2$ & 
$3,3,1$ & $1,2,2$ & $3,3,2$ & $3,6,7$ & 
$6, 4, 1$ &  $8,6,2$ & $3,3,1$ 
& $1,3,3$ \\ &&&&&&&&&&&& \\ \hline
LT & $1,3,3$ & $8,7,4$ & $1,3,3$  & $10,8,5$ & 
$2,2,1$ & $3,3,1$ & $3,3,2$ & $1,5,7$ & 
$3,3,1$ & $7,6,3$ & $2,3,2$ 
& $5,4,3$ \\ &&&&&&&&&&&& \\  \hline
\end{tabular}
\label{PARNN_parameters}
\end{table*}

%\begin{landscape}
\begin{table*}
\centering 
\caption{Short-term forecast performance comparison in terms of MASE, RMSE, and SMAPE of proposed PARNN model with statistical, machine learning, and deep learning forecasters. Mean values and (standard deviations) of 10 repetitions are reported in the table and the best-performing models are \underline{\textbf{highlighted}}.} 
%SHORT TERM COMPARISON IN TERMS OF MASE, RMSE AND SMAPE OF PROPOSED PARNN WITH STATISTICAL, MACHINE LEARNING, AND DEEP LEARNING FORECASTERS}
\tiny
    \begin{tabular}{|c|c|c|c|c|c|c|c|c|c|c|c|c|c|c|c|}%{|p{0.057\textwidth}|p{0.042\textwidth}|p{0.04\textwidth}|p{0.044\textwidth}|p{0.04\textwidth}|p{0.035\textwidth}|p{0.060\textwidth}|p{0.04\textwidth}|p{0.04\textwidth}|p{0.04\textwidth}|p{0.04\textwidth}|p{0.048\textwidth}|p{0.04\textwidth}|p{0.04\textwidth}|p{0.048\textwidth}|}
    
    \hline
        Dataset & Metric & 

        RWD	 & 	ETS	 & 	ARIMA	& 	TBATS	 & 	MLP	 & 	ARNN & Hybrid-1  & 	Hybrid-2  & NBeats	 & DeepAR	 & 	TCN	 & 	Transfor- & 	Hybrid-3  & \textcolor{blue}{PARNN}\\ 
        & &\cite{entorf1997random} &  \cite{winters1960forecasting} & \cite{box1970distribution}  & \cite{de2011forecasting} & \cite{rumelhart1986learning} & \cite{faraway1998time}  & \cite{zhang2003time} & \cite{chakraborty2019forecasting} &  \cite{oreshkin2019n} &  \cite{salinas2020deepar}  &  \cite{chen2020probabilistic}  & mers \cite{wu2020deep} & \cite{dave2021forecasting} & \textcolor{blue}{(Proposed)} \\\hline
         \multirow{5}{*}{AMZN}   
                                  & \textit{MASE} & 3.352	&	3.648	&	3.679 	&	3.087	&	\underline{\textbf{2.527}}	&	2.566	&	3.639 &	3.631	&	2.943	&	8.428 &	8.296	&	3.889	&	3.705	&	3.568 \\ 

                                  & &  (0)	&  (0)	&  (0)	&  (0)	& (0.04)	& (0.01)	& (0.001)	& (0.02)	& (0.43)	& (11.3)	& (15.3)	& (2.07)	& (0.11)	& (0.01) \\ 
                                 
                                  & \textit{RMSE} & 	12.00 &	13.17  &	13.26  &	10.76  &		\underline{\textbf{8.444}} &	8.502 &	13.17 &	13.14 &	10.75 &	26.52 &	26.94 &	13.26 &	13.56 &	10.82 \\

            Stock              && 	(0)	&	(0)	&	(0)	&	(0)	&	(0.14)	&	(0.03)	&	(0.003)	&		(0.07)	& (1.41)	&	(33.1)	&	(44.6)	&	(6.13)	&	(0.40)	& (0.03) \\
                                  & \textit{SMAPE} &	6.514	&	7.113	&	7.177	&	5.985	&	\underline{\textbf{4.896}}	&	4.973 &	7.098 &		7.080	&	5.710 	&	21.08 &	25.56 &	7.676 &	7.236 &	6.308 \\
                                  
                                  &&	(0)	& (0)	&  (0)	&  (0)	& (0.08)	& (0.02)	& (0.002)	& (0.05)	& (0.82) 	&	(33.1)	& (59.5)	& (4.56)	& (0.23)	& (0.02) \\ \hline
        \multirow{5}{*}{GOOG}   
                                & \textit{MASE} & 3.034 &	2.671 &	3.074 &	4.017 &	1.869 &	2.149 &	3.067 &	3.078	&	2.931 &	65.95 &	11.81 &	27.79 &	3.038 &	\underline{\textbf{1.437}} \\

                                && (0)	&	(0)	&	(0)	&	(0)	&	(0.07)	&	(0.01)	& (0.001)	&	(0.004)	& (1.59)	&	(1.22)	&	(20.8)	&	(19.4)	&	(0.06)	&	(0.02) \\

                                & \textit{RMSE} & 137.5 &	119.3 &	139.3	&	186.4 &	84.09	&	94.75 &	139.1 & 139.4 &	136.4 &	2620 &	507.1 &	1109 &	137.6 &	\underline{\textbf{68.85}} \\

                Stock                & & (0)	&	(0)	&	(0)	&	(0)	&	(2.83)	&	(0.46)	&	(0.03)	&	(0.15)	&	(74.2)	&	(48.5)	&	(830)	&	(768)	&	(2.76)	&	(0.34) \\
               
                                & \textit{SMAPE} & 4.506 	&	3.957 	&	4.567 	&	6.019 &	2.768 &	3.178 &	4.557 &	4.573 &	4.391 &	188.3 &	27.92 &	60.31 &	4.513 &	\underline{\textbf{2.135}} \\
                                && (0)	&	(0)	&	(0)	&	(0)	&	(0.11)	&	(0.02)	&	(0.001)	&	(0.006)	&	(2.50)	&	(6.79)	&	(59.9)	&	(56.2)	&	(0.09)	&	(0.03) \\\hline
        \multirow{5}{*}{MSFT}   
                                 & \textit{MASE} & 2.759 &	3.213 &	2.683 &	3.198 &	2.159 &	2.594 &	2.684 &	2.682 &	2.266 &	20.26 &	8.649 &	11.77 &	2.629 &	\underline{\textbf{1.385}} \\

                                 && (0)	&	(0)	&	(0)	&	(0)	& (0.08)	&	(1.03)	&	(0.001)	& (0.001)	& (0.64)	& (17.9)	& (18.5)	& (6.73)	& (0.06)	&	(0.07) \\                               
                                 & \textit{RMSE} & 15.02 &	17.84 &	14.61 &	17.76 &	11.64 &	14.12 &	14.61 &	14.61 &	12.66 &	98.65 &	44.55 &	57.54 &	14.54 &	\underline{\textbf{8.302}}\\

                   Stock              && (0)	&	(0)	&	(0)	&	(0)	&	(0.45)	&	(5.49)	&	(0.004)	&	(0.003)	& (3.57)	&	(85.2)	& (87.9)	& (32.0)	& (0.32)	&	(0.22) \\
                   
                                 & \textit{SMAPE} & 4.536	&	5.305	&	4.408	&	5.279 &	3.536 &	4.284 &	4.409 &	4.407 &	3.719 &	47.19 &	23.52 &	21.98 &	4.319 &	\underline{\textbf{2.282}} \\
                            && (0)	&	(0)	&	(0)	&	(0)	& (0.13)	&	(1.78)	&	(0.002)	&	(0.001)	&	(1.07)	&	(49.7)	&	(59.9)	&	(15.9)	&	(0.10)	&	(0.11) \\ \hline
        \multirow{6}{*}{Births}   
                                & \textit{MASE} &	2.416	&	1.370	&	1.210	&	1.222	&	1.487	&	1.552	&	1.893	&	1.173	&	2.117	&	11.35	&	3.520	&	1.625	&	1.205	&	\underline{\textbf{0.940}} \\
                                &&	(0)	&	(0)	&	(0)	&	(0)	&	(0.01)	&	(0.02)	&	(0.97)	&	(0.01)	&	(0.05)	&	(2.83)	&	(3.16)	&	(0.003)	&	(0.004)	&	(0.29) \\
                                & \textit{RMSE} & 2538	&	1376	&	1293	&	1286	&	1686	&	1700	&	1374	&	1283	&	2247	&	3648	&	3873	&	1671	&	1302	&	\underline{\textbf{1087}} \\ 
                                && 	(0)	&	(0)	&	(0)	&	(0)	&	(18.7)	&	(24.6)	&	(4.98)	&	(5.55)	&	(17.8)	&	(78.5)	&	(2963)	&	(3.40)	&	(2.93)	&	(309) \\
                                & \textit{SMAPE} & 22.41	&	12.05	&	10.59	&	10.72	&	14.98	&	14.05	&	10.67	& 10.24	&	20.16	&	19.1	&	37.09	&	14.42	&	10.55	&	\underline{\textbf{8.255}} \\
                                && (0)	&	(0)	&	(0)	&	(0)	&	(0.19)	&	(0.18)	&	(0.07)	&	(0.04)	&	(7.83)	&	(23.9)	&	(55.6)	&	(0.03)	&	(0.03)	&	(2.48) \\ \hline        
        \multirow{5}{*}{Colombia}   
                                & \textit{MASE} &  2.129	&	2.096	&	1.754	&	1.388	&	2.214	&	1.868	&	1.751	&	1.683	&	\underline{\textbf{1.349}}	&	7.467	&	4.113	&	6.851	&	1.737	&	2.030 \\
                                &&	(0)	&	(0)	&	(0)	&	(0)	&	(0.06)	&	(0.34)	&	(0.003)	&	(0.08)	&	(0.49)	&	(1.68)	&	(2.12)	&	(5.23)	&	(0.08)	&	($3 E^{-15}$) \\
                                & \textit{RMSE} &  254.3	&	250.4	&	227.3	&	196.3 &	260.0 &	199.5	&	227.1	&	227.5 &	\underline{\textbf{188.2}}	&	699.0	&	454.6	&	658.4	&	227.4	&	245.9 \\
                    Dengue            && (0)	&	(0)	&	(0)	&	(0)	&	(5.13)	&	(41.9)	&	(0.16)	&	(6.72)	&	(36.3)	&	(150)	&	(180)	&	(471)	&	(5.46)	&	($2 E^{-13}$)\\
                                & \textit{SMAPE} &  19.76	&	19.50	&	16.84	&	\underline{\textbf{13.90}}	&	20.39	&	20.01	&	16.81	&	16.24	&	14.13	&	110.2	&	44.11	&	51.95	&	16.69	&	19.01 \\

                                && (0)	& (0)	& (0)	& (0)	&	(0.44)	&	(4.03)	&	(0.02)	&	(0.61)	&	(5.54)	&	(39.4)	&	(43.1)	&	(33.7)	&	(0.60)	&	($2 E^{-14}$) \\ \hline
        \multirow{5}{*}{Colombia}   
                                & \textit{MASE} & 1.475	&	1.481	&	1.472	&	1.617	&	1.424	&	1.546	&	1.488	&	1.463	&	1.670	&	6.470	&	3.945	&	3.099	&	1.478	&	\underline{\textbf{1.093}} \\
                                && (0)	&	(0)	&	(0)	&	(0)	&	(0.04)	&	(0.09)	&	(0.001)	&	(0.03)	&	(0.19)	&	(0.86)	&	(1.69)	&	(1.04)	&	(0.03)	&	(0.02)\\
                                
                                & \textit{RMSE} & 258.8	&	276.4	&	266.2	&	306.3	&	259.0	&	288.0	&	265.9	&	263.0	&	237.8	&	822.9	&	588.4	&	446.3	&	259.4	&	\underline{\textbf{181.8}} \\
               Malaria      &&	(0)	&	(0)	&	(0)	&	(0)	&	(14.9)	&	(22.3)	&	(0.10)	&	(3.82)	&	(14.7)	&	(102)	&	(246)	&	(121)	&	(4.01)	&	(2.01) \\
            
                                & \textit{SMAPE} & 21.22	&	21.12	&	21.11	&	22.37	&	20.59	&	21.74	&	21.31	&	21.05	&	24.45	&	132.4	&	45.94	&	41.70 	&	21.24 	&	\underline{\textbf{16.59}} \\
                                &&	(0)	&	(0)	&	(0)	&	(0)	&	(0.39)	&	(0.86)	&	(0.004)	&	(0.29)	&	(2.81)	&	(31.6)	&	(10.7)	&	(23.5)	&	(0.34)	&	(0.36) \\ \hline
        \multirow{5}{*}{Venezuela}   
                                & \textit{MASE} &	4.248	&	3.877	&	4.129	&	4.198	&	4.058	&	4.059	&	4.114	&	4.128	&	4.839	&	6.437	&	4.528	&	4.061	&	4.089	&	\underline{\textbf{3.875}} \\

                                &&	(0)	&	(0)	&	(0)	&	(0)	& (0.03)	&	(0.04)	& (0.002)	& (0.001)	& (0.45)	& (1.03)	& (1.22)	& (1.32)	& (0.05)	& (0.04) \\
                                
                                & \textit{RMSE}	&	813.2	&	\underline{\textbf{742.3}}	&	794.8	&	804.6 &	778.3 &	777.3 &	792.8 &	794.7 &	926.4 &	1135 &	853.8 &	784.5 &	789.8 &	753.9  \\

            Dengue                    &&	(0)	&	(0)	&	(0)	&	(0)	&	(5.46)	&	(7.49)	&	(0.23)	&	(0.14)	&	(72.9)	&	(156)	&	(209)	&	(209)	&	(6.62)	&	(6.73) \\
            
                                & \textit{SMAPE} &	62.99 &	55.66 &	60.48 &	61.95 &	59.14 &	59.21 &	60.18 &	60.47 &	76.32 &	125.3 &	75.19 &	62.19 &	59.65	&	\underline{\textbf{55.61}} \\
                                                              
                                &&	(0)	&	(0)	&	(0)	&	(0)	&	(0.60)	&	(0.86)	& (0.03)	& (0.02)	& (10.7)	& (35.1)	& (26.1)	& (29.0)	& (0.99)	& (0.67) \\\hline

        \multirow{5}{*}{Venezuela}   
                                & \textit{MASE} &	0.798	&	\underline{\textbf{0.797}}	&	0.824	&	0.801	&	0.921	&	1.071 &	0.802 &	0.815 &	1.292 &	10.17 &	3.139 &	6.271 &	0.832 	&	0.825  \\	

                                && (0)	& (0)	& (0)	&	(0)	&	(0.05)	& (0.02)	& (0.001)	& (0.001)	& (0.42)	& (1.22)	& (1.49)	& (1.41)	& (0.04)	& (0.02) \\	 
                                & \textit{RMSE} & 121.0 & \underline{\textbf{120.6}} &	121.2 &	120.7 &	129.4 &	152.1 &	120.8 &	120.9	&	193.9 &	1337 &	501.9 &	831.5 &	122.9 &	138.8 \\ 

            Malaria         && (0)	&	(0)	&	(0)	&	(0)	&	(5.34)	&	(5.13)	& (0.01)	& (0.04)	& (63.6)	&	(159)	& (212)	& (183)	& (2.33)	& (3.76) \\ 
                  
                                & \textit{SMAPE} & 6.519 &	\underline{\textbf{6.510}} &	6.731  &	6.536 &	7.528 &	8.783 &	6.552 &	6.653 &	10.86 &	140.9	&	27.96 &	69.13	&	6.791 &	6.725 \\ 

                                && (0)	&	(0)	& (0)	& (0)	& (0.40)	& (0.20)	& (0.01)	& (0.01)	& (3.99)	& (28.8)	& (20.8)	& (24.7)	& (0.29)	& (0.16)  \\ \hline
      \multirow{5}{*}{US-EPU}   
                                & \textit{MASE} & 0.799	&	0.753	&	0.849	&	0.823	&	0.615	&	0.637	&	0.871	&	0.868	&	2.128	&	2.139	&	2.745	&	1.332	&	0.970	&	\underline{\textbf{0.534}} \\
                                && (0)	&	(0)	&	(0)	&	(0)	&	(0.03)	&	(0.04)	&	(0.01)	&	(0.003)	&	(0.52)	&	(1.97)	&	(1.88)	&	(1.37)	&	(0.11)	&	(0.06) \\
                                & \textit{RMSE} & 17.24	&	15.88	&	19.75	&	17.90	&	13.73	&	13.84	&	20.40	&	20.41	&	50.29	&	41.42	&	55.65	&	27.17	&	22.49	&	\underline{\textbf{12.91}} \\
                    Index       && (0)	&	(0)	&	(0)	&	(0)	&	(0.55)	&	(0.45)	&	(0.09)	&	(0.04)	&	(10.7)	&	(34.6)	&	(34.6)	&	(24.1)	&	(2.19)	&	(0.48) \\ 
                                & \textit{SMAPE} & 10.14	&	9.523	&	10.49	&	10.45	&	7.773	&	8.040	&	10.74	&	10.70	&	24.32	&	35.46	&	43.46	&	19.96	&	11.84	&	\underline{\textbf{6.671}} \\
                                && (0)	&	(0)	&	(0)	&	(0)	&	(0.35)	&	(0.47)	&	(0.05)	&	(0.03)	&	(5.71)	&	(40.8)	&	(47.8)	&	(26.3)	&	(1.27)	&	(0.67) \\ \hline
        \multirow{5}{*}{UK unem-}   
                                & \textit{MASE} & 2.674	&	0.957	&	0.720	&	0.951	&	0.763	&	0.716	&	0.720	&	0.719	&	1.427	&	2.760	&	5.187	&	2.426	&	0.737	&	\underline{\textbf{0.493}} \\
                                &&	(0)	&	(0)	&	(0)	&	(0)	&	(0.06)	&	(0.01)	&	(0.002)	&	(0.001)	&	(0.52)	&	(1.18)	&	(3.33)	&	(1.61)	&	(0.03)	&	 (0.003) \\
                                & \textit{RMSE} &  0.123	&	0.048	&	0.031	&	0.047	&	0.036	&	0.036	&	0.031	&	0.031	&	0.067	&	0.128	&	0.251	&	0.106	&	0.033	&	\underline{\textbf{0.023}} \\
           ployment             &&	(0)	&	(0)	&	(0)	&	(0)	&	(0.004)	&	(0.003)	&	(0.001)	&	(0.001)	&	(0.03)	&	(0.05)	&	(0.16)	&	(0.07)	&	(0.003)	&	(0.001) \\
                                & \textit{SMAPE} & 2.211	&	0.798	&	0.598	&	0.793	&	0.636	&	0.598	&	0.601	&	0.600	&	1.187	&	2.284	&	4.265	&	1.998	&	0.614	&	\underline{\textbf{0.411}} \\
                                && (0)	&	(0)	&	(0)	&	(0)	&	(0.05)	&	(0.01)	&	(0.002)	&	(0.001)	&	(0.43)	&	(0.97)	&	(2.56)	&	(1.31)	&	(0.02)	&	(0.002) \\ \hline 
     
     \multirow{5}{*}{Russia}   
                                & \textit{MASE} & 1.865	&	0.928	&	0.878	&\underline{\textbf{0.645}}	&	0.748	&	1.078	&	0.856	&	0.818	&	2.141	&	8.808	&	10.37	&	10.46	&	1.392	&	0.875 \\
                                && (0)	&	(0)	&	(0)	&	(0)	&	(0.07)	&	(0.17)	&	(0.004)	&	(0.01)	&	(0.93)	&	(14.1)	&	(11.6)	&	(9.68)	&	(0.24)	&	(0.02) \\
                                & \textit{RMSE} & 2.078	&	0.981	&	1.129	&	\underline{\textbf{0.792}}	&	0.860	&	1.266	&	1.113	&	1.084	&	2.431	&	8.811	&	11.74	&	10.35	&	1.741	&	0.973 \\
             Exchange                   && (0)	&	(0)	&	(0)	&	(0)	&	(0.08)	&	(0.18)	&	(0.003)	&	(0.002)	&	(1.015)	&	(13.8)	&	(12.5)	&	(9.54)	&	(0.27)	&	(0.03) \\
            
                                & \textit{SMAPE} & 2.497	&	1.258	&	1.191	&	\underline{\textbf{0.874}}	&	1.012	&	1.465	&	1.161	&	1.109	&	2.849	&	15.14	&	17.15	&	16.41	&	1.886	&	1.184 \\
                                && (0)	&	(0)	&	(0)	&	(0)	&	(0.09)	&	(0.23)	&	(0.01)	&	(0.01)	&	(1.22)	&	(27.2)	&	(26.1)	&	(18.5)	&	(0.32)	&	(0.02) \\ \hline
       \multirow{6}{*}{Tourism}   
                                & \textit{MASE} & \underline{\textbf{0.647}}	&	0.751	&	0.716	&	0.842	&	0.956	&	0.751	&	0.701	&	0.708	&	1.118	&	7.022	&	1.672	&	6.328	&	0.710	&	0.917 \\
                                && (0)	&	(0)	&	(0)	&	(0)	&	(0.07)	&	(0.01)	&	(0.003)	&	(0.001)	&	(0.19)	&	(0.09)	&	(1.89)	&	(0.64)	&	(0.003)	&	(0.24) \\
                                & \textit{RMSE} & 	86.71	&	\underline{\textbf{80.46}}	&	83.62	&	83.59	&	94.63	&	104.4	&	84.04	&	84.06	&	109.8	&	652.5	&	172.1	&	589.4	&	83.82	&	98.19 \\
                                && (0)	&	(0)	&	(0)	&	(0)	&	(5.63)	&	(0.59)	&	(0.06)	&	(0.01)	&	(19.6)	&	(8.51)	&	(175)	&	(58.3)	&	(0.31)	&	(29.4) \\
                                & \textit{MASE} & \underline{\textbf{8.674}}	&	10.19	&	9.679	&	11.44	&	12.94	&	10.22	&	9.460	&	9.563	&	15.06	&	191.5	&	31.05	&	159.3	&	9.590	&	12.56 \\
                                &&	(0)	&	(0)	&	(0)	&	(0)	&	(0.86)	&	(0.08)	&	(0.04)	&	(0.01)	&	(2.40)	&	(5.04)	&	(51.2)	&	(28.3)	&	(0.03)	&	(3.23) \\ \hline
    \end{tabular}\label{S-finaltable}
\end{table*}
% \end{landscape}

% \begin{landscape}
    
\begin{table*}
\centering \caption{Medium-term forecast performance comparison in terms of MASE, RMSE, and SMAPE of proposed PARNN model with statistical, machine learning, and deep learning forecasters. Mean values and (standard deviations) of 10 repetitions are reported in the table and the best-performing models are \underline{\textbf{highlighted}}.}

%Medium TERM COMPARISON IN TERMS OF RMSE, MASE, AND SMAPE OF PROPOSED PARNN WITH STATISTICAL, MACHINE LEARNING, AND DEEP LEARNING FORECASTERS }
\tiny
    \begin{tabular}{|c|c|c|c|c|c|c|c|c|c|c|c|c|c|c|c|}%{|p{0.057\textwidth}|p{0.042\textwidth}|p{0.04\textwidth}|p{0.044\textwidth}|p{0.04\textwidth}|p{0.035\textwidth}|p{0.060\textwidth}|p{0.04\textwidth}|p{0.04\textwidth}|p{0.04\textwidth}|p{0.04\textwidth}|p{0.048\textwidth}|p{0.04\textwidth}|p{0.04\textwidth}|p{0.048\textwidth}|}
    
    \hline
        Dataset & Metric & 

        RWD	 & 	ETS	 & 	ARIMA	& 	TBATS	 & 	MLP	 & 	ARNN & Hybrid-1  & 	Hybrid-2  & NBeats	 & DeepAR	 & 	TCN	 & 	Transfor- & 	Hybrid-3  & \textcolor{blue}{PARNN}\\ 
        & &\cite{entorf1997random} &  \cite{winters1960forecasting} & \cite{box1970distribution}  & \cite{de2011forecasting} & \cite{rumelhart1986learning} & \cite{faraway1998time}  & \cite{zhang2003time} & \cite{chakraborty2019forecasting} &  \cite{oreshkin2019n} &  \cite{salinas2020deepar}  &  \cite{chen2020probabilistic}  & mers \cite{wu2020deep} & \cite{dave2021forecasting} & \textcolor{blue}{(Proposed)} \\\hline
        \multirow{5}{*}{AMZN}   
                            & \textit{MASE} &	2.987	&	2.552	&	2.550	&	3.420	&	3.472	&	3.313	&	2.577	&	2.588	&	4.098	&	8.278	&	9.096	&	4.071	&	2.610	&	\underline{\textbf{2.440}} \\
                            &&	(0)	&	(0)	&	(0)	&	(0)	&	(0.13)	&	(0.01)	&	(0.001)	&		(0.01)	&	(0.37)	&	(9.37)	&	(13.8)	&	(3.09)	&	(0.10)	&	($3 E^{-4}$) \\
                            & \textit{RMSE} &  12.98	&	10.93	&	10.93	&	14.35	&	14.53	&	14.01	&	11.03	&	11.07	&	16.53	&	30.74	&	38.41	&	16.44	&	11.19	&	\underline{\textbf{10.40}} \\	
                        Stock    &&	(0)	&	(0)	&	(0)	&	(0)	&	(0.41)	&	(0.02)	&	(0.004)	&	(0.03)	&	(1.19)	&	(32.3)	&	(56.9)	&	(10.4)	&	(0.37)	&	($1 E^{-3}$) 	\\ 
            
                            & \textit{SMAPE} &  6.816	&	5.867	&	5.864	&	7.749	&	7.860	&	7.520	&	5.924	&	5.947	&	9.186	&	24.29	&	23.30	&	9.820	&	5.993	&	\underline{\textbf{5.620}} \\
                            &&	(0)	&	(0)	&	(0)	&	(0)	&	(0.26)	&	(0.01)	&	(0.002) &	(0.02)	&	(0.77)	&	(33.8)	&	(40.9)	&	(8.74)	&	(0.22)	&	($8 E^{-4}$) \\ \hline
       \multirow{5}{*}{GOOG}   
                            & \textit{MASE} & 2.793	&	2.843	&	2.870	&	1.804	&	2.697	&	2.410	&	2.869	&	2.867	&	5.364	&	55.12	&	12.05	&	25.43	&	2.874	&	\underline{\textbf{1.758}} \\
                            &&	(0)	&	(0)	&	(0)	&	(0)	&	(0.09)	&	(0.004)	&	(0.004)	&	(0.002)	&	(0.61)	&	(0.96)	&	(17.9)	&	(17.0)	&	(0.03)	&	(0.01) \\
                            
                            & \textit{RMSE} & 156.8	&	159.1	&	160.3	&	103.5 &	152.5	&	139.5	&	160.2	&	160.2	&	275.4	&	2626	&	673.9	&	1217 &	160.3	&	\underline{\textbf{100.9}} \\
            Stock                &&	(0)	&	(0)	&	(0)	&	(0)	&	(4.24)	&	(0.19)	&	(0.18)	&	     (0.06)	&	(29.2)	&	(45.4)	&	(996)	&	(806)	&	(1.48)	&	(0.31)\\
            
                       & \textit{SMAPE} & 	4.856	&	4.940	&	4.985	&	3.175	&	4.699	&	4.216	&	4.984	&	4.981	&	9.065	&	189.1	&	23.67	&	68.02	&	4.992	&	\underline{\textbf{3.093}}\\
                            && (0)	&	(0)	&	(0)	&	(0)	&	(0.16)	&	(0.01)	&	(0.01)	&	(0.003)	&	(0.99)	&	(6.38)	&	(41.5)	&	(59.8)	&	(0.06)	&	(0.03) \\ \hline
      \multirow{5}{*}{MSFT}   
                            & \textit{MASE} & 2.828	&	\underline{\textbf{1.684}}	&	2.968	&	1.713	&	3.774	&	3.084	&	2.968	&	2.968	&	7.013	&	20.49	&	11.63	&	12.06	&	2.952	&	1.845 \\
                            && (0)	&	(0)	&	(0)	&	(0)	&	(0.14)	&	(0.01)	&	(0.001)	&		(0.001)	&	(0.58)	&	(17.6)	&	(18.9)	&	(8.11)	&	(0.11)	&	(0.03)\\
                            & \textit{RMSE} & 17.38 & \underline{\textbf{10.94}}	&	18.08	&	11.11	&	22.37	&	18.84	&	18.08	&	18.08	&	37.99	&	104.4	&	70.89	&	61.68	&	17.84	&	11.77 \\
            Stock                && (0)	&	(0)	&	(0)	&	(0)	&	(0.73)	&	(0.04)	&	(0.01)	&	(0.004)	&	(2.67)	&	(87.9)	&	(112)	&	(40.7)	&	(0.59)	&	(0.16) \\
            
                            & \textit{SMAPE} &  4.740	&	\underline{\textbf{2.867}}	&	4.966	&	2.916	&	6.257	&	5.155	&	4.966	&	4.966	&	11.27	&	50.36	&	22.13	&	24.29	&	4.942	&	3.136 \\
                            && 	(0)	&	(0)	&	(0)	&	(0)	&	(0.23)	&	(0.01)	&	(0.002)		&	(0.001)	&	(0.87)	&	(51.3)	&	(41.5)	&	(21.4)	&	(0.18)	&	(0.05) \\ \hline
      \multirow{6}{*}{Births}   
                            & \textit{MASE} & 	2.447	&	1.362	&	1.352	&	1.344	& 2.169		&	2.180	&	1.254	&	1.276	&	6.439	&	11.25	&	9.102	&	1.571	&	1.313	&	\underline{\textbf{0.636}}\\
                            &&	(0)	&	(0)	&	(0)	&	(0)	&	(0.01)	&	(0.001)	&	(0.01)	&	(0.01)	&	(0.56)	&	(13.8)	&	(11.7)	&	(0.002)	&	(0.03)	&	(0.05) \\
                            & \textit{RMSE} &	2553	&	1354	&	1376	&	1369	&	2229	&	2227	&	1365	&	1349	&	7417	&	10$E^3$	&	10$E^3$	&	1597	&	1352	&	\underline{\textbf{760.8}} \\
                            && 	(0)	&	(0)	&	(0)	&	(0)	&	(0.001)	&	(0.01)	&	(4.87)	&	(5.07)	&	(83.9)	&	(97.4)	&	(14$E^3$)	&	(1.98)	&	(20.6)	&	(94.1) \\
                            & \textit{SMAPE} &	23.00	&	12.09	&	12.01	&	11.93	& 20.53		&	20.59	&	11.25	&	11.30	&	39.54	&	191.2	&	64.20	&	14.05	&	11.64	&	\underline{\textbf{5.610}} \\
                            && (0)	&	(0)	&	(0)	&	(0)	&	(0.1)	&	(0.01)	&	(0.57)	&	(0.06)	&	(1.95)	&	(11.7)	&	(68.2)	&	(0.02)	&	(0.24)	&	(0.42) \\ \hline
        \multirow{5}{*}{Colombia}   
                                & \textit{MASE} & 8.720	&	8.718	&	9.069	&	8.613	&	10.33	&	10.69	&	9.075	&	9.098	&	10.17	&	10.09	&	56.87	&	6.197	&	9.118	&	\underline{\textbf{1.762}} \\
                                && (0)	&	(0)	&	(0)	&	(0)	&	(0.13)	&	(0.59)	&	(0.003)	&	(0.08)	&	(0.99)	&	(1.48)	&	(68.5)	&	(2.76)	&	(0.12)	&	(0.23) \\
                                & \textit{RMSE}  & 	922.4	&	917.2	&	949.8	&	906.0	&	1083	&	1122	&	953.0	&	955.2	&	1089	&	1043	&	7421	&	680.4	&	957.5	&	\underline{\textbf{222.6}} \\
            Dengue                    && 	(0)	&	(0)	&	(0)	&	(0)	&	(14.7)	&	(63.6)	&	(0.39)	&	(6.16)	&	(136)	&	(136)	&	(9200)	&	(259)	&	(13.3)	&	(20.1) \\
            
                                & \textit{SMAPE}  & 53.29	&	53.33	&	54.79	&	52.90	&	59.58	&	60.87	&	54.79	&	54.86	&	58.36	&	126.1	&	109.6	&	49.47	&	54.96	&	\underline{\textbf{14.76}} \\
                                &&	(0)	&	(0)	&	(0)	&	(0)	&	(0.47)	&	(2.11)	&	(0.01)	&	(0.32)	&	(3.25)	&	(32.7)	&	(36.9)	&	(32.4)	&	(0.49)	&	(1.76) \\ \hline
       \multirow{5}{*}{Colombia}
                                & \textit{MASE} & 4.436	&	4.997	&	4.661	&	4.916	&	3.330	&	2.733	&	4.652	&	4.631	&	4.599	&	8.834	&	10.87	&	3.911	&	4.481	&	\underline{\textbf{2.543}}	\\
                                && 	(0)	&	(0)	&	(0)	&	(0)	&	(0.23)	&	(0.24)	&	(0.01)	&	(0.04)	&	(1.29)	&	(0.91)	&	(4.24)	&	(1.94)	&	(0.04)	&	(0.002)	\\
                                & \textit{RMSE} &	591.1	&	655.3	&	617.0	&	646.8	&	444.8	&	404.8	&	615.3	&	610.9 &	637.1 	&	1059	&	1507	&	519.7	&	595.6	&	\underline{\textbf{361.7}} \\	
            Malaria                    && (0)	&	(0)	&	(0)	&	(0)	&	(26.4)	&	(27.7)	&	(0.56)	&	(5.64)	&	(156)	&	(99.1)	&	(526)	&	(211)	&	(4.24)	&	(0.15)	\\
            
                                & \textit{SMAPE} & 38.59 &	42.03	&	39.99	&	41.54	&	31.28	&	26.55	&	39.94	&	39.81	&	38.53	&	143.8	&	71.52	&	42.79	&	38.87	&	\underline{\textbf{25.26}}\\
                                &&	(0)	&	(0)	&	(0)	&	(0)	&	(1.63)	&	(1.89)	&	(0.03)	&	(0.26)	&	(8.75)	&	(27.5)	&	(27.4)	&	(31.9)	&	(0.25)	&	(0.02)	\\ \hline
      \multirow{5}{*}{Venezuela}   
                                & \textit{MASE} &	3.381	&	3.881	&	3.179	&	3.347	&	3.223	&	3.167	&	3.183	&	3.180	&	3.833	&	5.536	&	7.674	&	3.157	&	3.149	&	\underline{\textbf{2.628}} \\
                                && (0)	&	(0)	&	(0)	&	(0)	&	(0.01)	&	(0.01)	&	(0.001)	&	(0.001)	&	(0.78)	&	(1.19)	&	(3.35)	&	(0.98)	&	(0.02)	&	(0.001)
                                \\
                                & \textit{RMSE} &	513.5	&	588.0	&	486.6	&	508.6	&	492.6	&	485.2	&	487.3	&	486.8	&	604.5	&	875.7	&	1292	&	518.5	&	482.9	&	\underline{\textbf{447.7}} \\ 
            Dengue                    && 	(0)	&	(0)	&	(0) &	(0)	&	(1.95)	&	(1.06)	&	(0.07)	&	(0.05)	&	(133)	&	(138)	&	(564)	&	(144)	&	(2.76)	&	(0.01)\\
            
                                & \textit{SMAPE} &  41.89	&	46.25	&	39.95	&	41.58	&	40.42	&	39.87	&	39.99	&	39.97	&	48.40	&	107.8	&	76.20	&	44.28	&	39.65	&	\underline{\textbf{33.79}}\\
                                &&  (0)	&	(0)	&	(0)	&	(0)	&	(0.13)	&	(0.09)	&	(0.01)	&	(0.01)	&	(10.9)	&	(40.7)	&	(27.9)	&	(24.2)	&	(0.23)	&	(0.003) \\ \hline
        \multirow{5}{*}{Venezuela}   
                                & \textit{MASE} & 0.815	&	0.992	&	1.217	&	1.426	&	0.894	&	0.865	&	1.179	&	1.164	&	1.518	&	7.783	&	3.976	&	4.587	&	1.144	&	\underline{\textbf{0.768}} \\
                                && (0)	&	(0)	&	(0)	&	(0)	&	(0.05)	&	(0.08)	&	(0.003)	&	(0.04)	&	(0.63)	&	(0.92)	&	(2.63)	&	(1.31)	&	(0.05)	&	(0.01) \\
                                & \textit{RMSE} & \underline{\textbf{162.6}}	&	204.6 &	244.6	&	278.4	&	182.2	&	186.4	&	238.5	&	237.1	&	303.8	&	1332	&	803.2	&	794.8	&	232.8	&	171.2 \\
            Malaria                    && (0)	&	(0)	&	(0)	&	(0)	&	(12.6)	&	(13.4)	&	(0.36)	&	(5.56)	&	(100)	&	(154)	&	(525)	&	(219)	&	(7.84)	&	(2.44)\\
            
                                & \textit{SMAPE} & 8.677	&	10.68	&	13.32	&	15.86	&	9.562	&	9.236	&	12.87	&	12.69	&	17.39	&	141.6	&	51.77	&	65.97	&	12.46	&	\underline{\textbf{8.249}} \\ 
                                &&	(0)	&	(0)	&	(0)	&	(0)	&	(0.60)	&	(0.79)	&	(0.03)	&	(0.48)	&	(8.13)	&	(28.5)	&	(34.3)	&	(29.7)	&	(0.53)	&	(0.11) \\ \hline
        \multirow{5}{*}{US EPU}   
                                & \textit{MASE} & 4.939	&	4.716	&	4.590	&	2.264	&	3.927	&	5.076	&	4.729	&	4.712	&	6.132	&	2.147	&	6.125	&	\underline{\textbf{1.567}}	&	4.772	&	1.727 \\
                                &&	(0)	&	(0)	&	(0)	&	(0)	&	(0.36)	&	(1.00)	&	(0.004)	&	(0.003)	&	(1.04)	&	(1.74)	&	(3.81)	&	(1.33)	&	(0.22)	&	(0.27)\\
                                & \textit{RMSE} & 104.9	&	100.3	&	97.62	&	51.93	&	84.43	&	110.5	&	100.5	&	100.1	&	136.4	&	48.42	&	145.6	&	\underline{\textbf{39.42}}	&	101.3	&	40.84 \\
            Index                    && 	(0)	&	(0)	&	(0)	&	(0)	&	(7.18)	&	(20.8)	&	(0.08)	&	(0.05)	&	(26.0)	&	(35.2)	&	(82.7)	&	(25.8)	&	(4.48)	&	(6.51) \\
            
                                & \textit{SMAPE} & 52.22	&	50.49	&	49.49	&	27.82	&	43.90	&	52.56	&	50.59	&	50.45	&	59.06	&	39.58	&	65.66	&	26.63	&	50.91	&	\underline{\textbf{22.36}}\\
                                && (0)	&	(0)	&	(0)	&	(0)	&	(3.13)	&	(8.05)	&	(0.03)	&	(0.02)	&	(6.41)	&	(41.6)	&	(42.5)	&	(30.7)	&	(1.71)	&	(2.98) \\ \hline
        \multirow{5}{*}{UK unem-}   
                                & \textit{MASE} & 6.154	&	2.167	&	1.365	&	2.128	&	1.634	&	2.159	&	1.357	&	1.361	&	9.866	&	3.483	&	17.30	&	10.14	&	1.344	&	\underline{\textbf{0.906}} \\
                                &&	(0)	&	(0)	&	(0)	&	(0)	&	(0.33)	&	(0.20)	&	(0.01)	&	(0.002)	&	(2.82)	&	(3.59)	&	(3.72)	&	(4.33)	&	(0.09)	&	(0.01) \\
                                & \textit{RMSE} & 0.258	&	0.089	&	0.060	&	0.087	&	0.072	&	0.091	&	0.059	&	0.059	&	0.396	&	0.160	&	0.858	&	0.382	&	0.059	&	\underline{\textbf{0.038}} \\
            ployment                    &&	(0)	&	(0)	&	(0)	&	(0)	&	(0.01)	&	(0.01)	&	(0.001)	&	(0.001)	&	(0.11) &	(0.16)	&	(0.21)	&	(0.15)	&	(0.004)	&	(0.001) \\
            
                                & \textit{SMAPE} & 4.501	&	1.612	&	1.024	&	1.584	&	1.215	&	1.607	&	1.018	&	1.021	&	7.072	&	2.534	&	12.15	&	7.245	&	1.008	&	\underline{\textbf{0.671}}\\
                                && (0)	&	(0)	&	(0)	&	(0)	&	(0.25)	&	(0.15)	&	(0.01)	&	(0.002)	&	(1.94)	&	(2.54)	&	(2.96)	&	(2.98)	&	(0.07)	&	(0.01) \\ \hline
      
       \multirow{5}{*}{Russia}   
                                & \textit{MASE} & 4.816	&	4.172	&	3.632	&	2.926	&	2.527	&	5.567	&	3.631	&	3.642	&	12.07	&	13.06	&	23.79	&	13.89	&	3.399	&	\underline{\textbf{1.250}} \\
                                &&	(0)	&	(0)	&	(0)	&	(0)	&	(0.19)	&	(0.47)	&	(0.01)`	&	(0.004)	&	(8.85)	&	(13.7)	&	(36.1)	&	(10.7)	&	(0.09)	&	(0.08)\\
                                & \textit{RMSE} & 4.912	&	4.197	&	3.874	&	3.028	&	2.646	&	5.750	&	3.873	&	3.883	&	12.31	&	12.81	&	25.89	&	13.47	&	3.766	&	\underline{\textbf{1.492}} \\
        Exchange                        &&	(0)	&	(0)	&	(0)	&	(0)	&	(0.17)	&	(0.48)	&	(0.01)	&	(0.004)	&	(8.78)	&	(13.1)	&	(36.4)	&	(10.2)	&	(0.10)	&	(0.11)\\
        
                                & \textit{SMAPE} & 6.088	&	5.301	&	4.627	&	3.750	&	3.248	&	7.573	&	4.626	&	4.640	&	14.01	&	20.97	&	32.26	&	21.32	&	4.335	&	\underline{\textbf{1.626}} \\
                                &&	(0)	&	(0)	&	(0)	&	(0)	&	(0.24)	&	(0.66)	&	(0.01)	&	(0.01)	&	(9.47)	&	(26.7)	&	(49.9)	&	(20.8)	&	(0.12)	&	(0.12) \\ \hline
        \multirow{6}{*}{Tourism}   
                                & \textit{MASE} & 0.773	&	0.752	&	\underline{\textbf{0.727}}	&	0.762	&	0.984	&	1.057	&	0.737	&	0.759	&	1.024	&	9.141	&	5.407	&	8.257	&	0.729	&	1.073	\\
                                &&	(0)	&	(0)	&	(0)	&	(0)	&	(0.07)	&	(0.02)	&	(0.001)	&	(0.002)	&	(0.09)	&	(0.12)	&	(7.19)	&	(0.82)	&	(0.003)	&	(0.003) \\ 
                                & \textit{RMSE} & 83.21	&	79.75	&	\underline{\textbf{76.45}}	&	81.84	&	94.74	&	101.9	&	77.93	&	80.99	&	91.51	&	652.1	&	415.4	&	590.0	&	76.80	&	102.1 \\
                                &&	(0)	&	(0)	&	(0)	&	(0)	&	(5.25)	&	(1.17)	&	(0.07)	&	(0.12)	&	(8.42)	&	(8.71)	&	(533)	&	(57.9)	&	(0.16)	&	(0.13)	\\
                                & \textit{SMAPE} & 	8.104	&	7.863	&	\underline{\textbf{7.588}}	&	7.982	&	10.55	&	11.42	&	7.693	&	7.947	&	10.59	&	191.9	&	62.26	&	160.2	&	7.602	&	11.62 \\
                                &&	(0)	&	(0)	&	(0)	&	(0)	&	(0.84)	&	(0.20)	&	(0.01)	&	(0.01)	&	(0.98)	&	(5.13)	&	(83.4)	&	(27.9)	&	(0.03)	&	0.03 \\ \hline                                
    \end{tabular}\label{M-finaltable}
\end{table*}
% \end{landscape}

% \begin{landscape}

\begin{table*}
\centering \caption{Long-term forecast performance comparison in terms of MASE, RMSE, and SMAPE of proposed PARNN model with statistical, machine learning, and deep learning forecasters. Mean values and (standard deviations) of 10 repetitions are reported in the table and the best performing models are \underline{\textbf{highlighted}}.}

%LOng-term COMPARISON (IN TERMS OF RMSE, MASE, AND SMAPE) OF PROPOSED PARNN WITH STATISTICAL, MACHINE LEARNING, AND DEEP LEARNING FORECASTERS }
\tiny
    \begin{tabular}{|c|c|c|c|c|c|c|c|c|c|c|c|c|c|c|c|}%{|p{0.057\textwidth}|p{0.042\textwidth}|p{0.04\textwidth}|p{0.044\textwidth}|p{0.04\textwidth}|p{0.035\textwidth}|p{0.060\textwidth}|p{0.04\textwidth}|p{0.04\textwidth}|p{0.04\textwidth}|p{0.04\textwidth}|p{0.048\textwidth}|p{0.04\textwidth}|p{0.04\textwidth}|p{0.048\textwidth}|}
    
    \hline
        Dataset & Metric & 
        RWD	 & 	ETS	 & 	ARIMA	& 	TBATS	 & 	MLP	 & 	ARNN & Hybrid-1  & 	Hybrid-2  & NBeats	 & DeepAR	 & 	TCN	 & 	Transfor- & 	Hybrid-3  & \textcolor{blue}{PARNN}\\ 
        & &\cite{entorf1997random} &  \cite{winters1960forecasting} & \cite{box1970distribution}  & \cite{de2011forecasting} & \cite{rumelhart1986learning} & \cite{faraway1998time}  & \cite{zhang2003time} & \cite{chakraborty2019forecasting} &  \cite{oreshkin2019n} &  \cite{salinas2020deepar}  &  \cite{chen2020probabilistic}  & mers \cite{wu2020deep} & \cite{dave2021forecasting} & \textcolor{blue}{(Proposed)} \\\hline
        \multirow{5}{*}{AMZN}
                            & \textit{MASE} & 7.926	&	4.674	&	4.674	&	3.929	&	3.384	&	3.304	&	4.722	&	4.724	&	5.284	&	12.49	&	13.00	&	5.630	&	4.667	&	\underline{\textbf{2.641}}\\
                            &&	(0)	&	(0)	&	(0)	&	(0)	&	(0.11)	&	(0.004)	&	(0.002)		&	(0.01)	&	(0.71)	&	(13.8)	&	(16.6)	&	(6.12)	&	(0.19)	&	(0.74) \\
                            & \textit{RMSE} & 27.37	&	17.01	&	17.01	&	14.79	&	12.55	&	12.24	&	17.15	&	17.16	&	18.76	&	37.79	&	44.09	&	17.67	&	17.03	&	\underline{\textbf{9.974}} \\
            Stock                &&  (0)	&	(0)	&	(0)	&	(0)	&	(0.39)	&	(0.02)	&	(0.004)	&	(0.03)	&	(2.38)	&	(38.2)	&	(51.8)	&	(17.2)	&	(0.49)	&	(2.66) \\
            
                            & \textit{SMAPE} & 13.22	&	8.141	&	8.141	&	6.917	&	5.998	&	5.864	&	8.219	&	8.222	&	9.117	&	28.94	&	27.57	&	10.91	&	8.125	&	\underline{\textbf{4.714}}\\
                            &&	(0)	&	(0)	&	(0)	&	(0)	&	(0.18)	&	(0.01)	&	(0.002)	&	(0.02)	&	(1.14)	&	(38.4)	&	(46.5)	&	(13.8)	&	(0.32)	&	(1.25) \\ \hline

         \multirow{5}{*}{GOOG}
                            & \textit{MASE} & 10.49	&	10.39	&	10.32	&	5.413	&	8.960	&	2.869	&	10.30	&	10.34	&	11.25	&	70.52	&	18.24	&	36.81	&	10.31	&	\underline{\textbf{2.830}} \\
                            &&	(0)	&	(0)	&	(0)	&	(0)	&	(0.23)	&	(0.01)	&	(0.01)	&	(0.03)	&	(0.56)	&	(1.05)	&	(20.8)	&	(20.1)	&	(0.07)	&	(0.02) \\ 
                            & \textit{RMSE} & 476.5	&	472.1	&	469.9	&	253.8	&	410.2	&	132.9	&	469.3	&	470.5	&	521.5	&	2737	&	843.5	&	1435	&	469.9	&	\underline{\textbf{128.4}} \\
            Stock                &&	(0)	&	(0)	&	(0)	&	(0)	&	(10.1)	&	(0.79)	&	(0.17)	&	(0.54)	&	(24.1)	&	(40.6)	&	(897)	&	(775)	&	(3.29)	&	(0.02) \\
        
                            & \textit{SMAPE} &  13.55	&	13.43	&	13.34	&	7.331	&	11.73	&	\underline{\textbf{3.722}}	&	13.32	&	13.36	&	14.38	&	190.7	&	29.02	&	77.98	&	13.32	&	3.938 \\
                            && (0)	&	(0)	&	(0)	&	(0)	&	(0.27)	&	(0.02)	&	(0.01)	&	(0.04)	&	(0.65)	&	(5.55)	&	(45.3)	&	(58.4)	&	(0.09)	&	(0.02) \\ \hline
         \multirow{5}{*}{MSFT}
                            & \textit{MASE} & 10.02	&	9.865	&	9.947	&	5.549	&	10.98	&	4.741	&	9.950	&	9.950	&	7.792	&	29.46	&	15.31	&	18.78	&	9.849	&	\underline{\textbf{3.240}} \\
                            && 	(0)	&	(0)	&	(0)	&	(0)	&	(0.54)	&	(0.05)	&	(0.002)	&	(0.001)	&	(0.68)	&	(19.7)	&	(20.4)	&	(10.1)	&	(0.08)	&	(0.02) \\
                            & \textit{RMSE} & 54.51	&	53.76	&	54.20	&	31.09 &	59.36	&	26.73	&	54.21	&	54.21	&	43.20	&	134.1	&	81.25	&	86.97	&	53.74	&	\underline{\textbf{17.59}} \\
        Stock                    && 	(0)	&	(0)	&	(0)	&	(0)	&	(2.86)	&	(0.28)	&	(0.01)	&	(0.003)	&	(4.41)	&	(87.2)	&	(100)	&	(44.7)	&	(0.38)	&	(0.33)\\
        
                            & \textit{SMAPE} &  13.52	&	13.32	&	13.43	&	7.841	&	14.67	&	6.761	&	13.43	&	13.43	&	10.75	&	61.63	&	26.93	&	32.81	&	13.31	&	\underline{\textbf{4.660}}\\
                            && 	(0)	&	(0)	&	(0)	&	(0)	&	(0.65)	&	(0.07)	&	(0.002)	&	(0.001)	&	(0.83)	&	(50.9)	&	(47.1)	&	(23.9)	&	(0.09)	&	(0.03) \\ \hline           
    \multirow{6}{*}{Births}   
                            & \textit{MASE} & 	2.383	&	1.492	&	1.462	&	1.422	&	2.301	&	2.289	&	1.426	&	1.397	&	3.862	&	10.94	&	62$E^3$	&	1.539	&	1.424	&	\underline{\textbf{0.554}} \\
                            &&  (0)	&	(0)	&	(0)	&	(0)	&	(0.001)	&	(0.01)	&	(1.87)	&	(0.04)	&	(2.51)	&	(1.67)	&	(19$E^3$)	&	(0.01)	&	(0.03)	&	(0.003) \\
                            & \textit{RMSE}	&	2432	&	1480	&	1442	&	1404	&	2187	&	2199	&	1407	&	1427	&	4165	&	9877	&	15$E^6$	&	1540	&	1426	& \underline{\textbf{640.5}} \\
                            &&	(0)	&	(0)	&	(0)	&	(0)	&	(0.03)	&	(0.15)	&	(19.2)	&	(25.2)	&	(2859)	&	(2176)	&	(49$E^5$)	&	(5.35)	&	(19.4)	&	(2.42)\\
                            & \textit{SMAPE}	&	22.03	&	13.13	&	12.86	&	12.49	&	22.01	&	21.41	&	13.98	&	12.25	&	28.56	&	176.8	&	90.49	&	13.57	&	12.51	&	\underline{\textbf{4.832}} \\
                            &&	(0)	&	(0)	&	(0)	&	(0)	&	(0.12)	&	(0.35)	&	(0.67)	&	(0.35)	&	(13.4)	&	(42.8)	&	(69.2)	&	(0.04)	&	(0.22)	&	(0.03) \\ \hline

        \multirow{5}{*}{Colombia}   
                                & \textit{MASE} & 5.337	&	4.862	&	4.865	&	4.860	&	4.324	&	4.096	&	4.864	&	4.865	&	18.535	&	10.483	&	7.645	&	5.928	&	4.935	&	\underline{\textbf{3.228}} \\
                                &&	(0)	&	(0)	&	(0)	&	(0)	&	(0.09)	&	(0.17)	&	(0.01)	&	(0.003)	&	(4.84)	&	(0.85)	&	(3.47)	&	(2.16)	&	(0.03)	&	(0.04) \\
                                & \textit{RMSE} & 1111	&	997.2	&	998.3	&	997.5	&	845.3	&	840.1	&	997.8	&	998.5	&	3610	&	1914	&	1727	&	1125	&	1015	&	\underline{\textbf{611.0}} \\
                    Dengue            && 	(0)	&	(0)	&	(0)	&	(0)	&	(22.8)	&	(39.1)	&	(1.68)	&	(0.74)	&	(1094)	&	(127)	&	(788)	&	(405)	&	(7.17)	&	(7.98) \\
        
                                & \textit{SMAPE} & 43.40	&	40.98	&	40.99	&	40.96	&	37.99	&	36.42	&	40.99	&	40.99	&	83.74	&	146.6	&	59.81	&	61.39	&	41.37	&	\underline{\textbf{28.94}} \\
                                &&	(0)	&	(0)	&	(0)	&	(0)	&	(0.51)	&	(1.03)	&	(0.03)	&	(0.01)	&	(8.62)	&	(24.9)	&	(29.1)	&	(37.5)	&	(0.16)	&	(0.36)\\ \hline

        \multirow{5}{*}{Colombia}   
                                & \textit{MASE} & 3.765	&	3.501	&	3.547	&	3.519	&	3.837	&	3.661	&	3.559	&	3.575	&	3.565	&	8.319	&	5.533	&	4.101	&	3.617	&	\underline{\textbf{3.272}} \\
                                &&	(0)	&	(0)	&	(0)	&	(0)	&	(0.23)	&	(0.19)	&	(0.001)	&	(0.01)	&	(0.49)	&	(0.54)	&	(1.83)	&	(1.65)	&	(0.04) &	(0.01) \\
                                & \textit{RMSE} & 850.5	&	798.6	&	810.3	&	803.0	&	849.4	&	811.7	&	813.2	&	816.4	&	759.9	&	1623	&	1231	&	880.4	&	826.3	&	\underline{\textbf{724.5}} \\
                    Malaria            &&	(0)	&	(0)	&	(0)	&	(0)	&	(33.8)	&	(32.0)	&	(0.24)	&	(3.29)	&	(86.9)	&	(91.9)	&	(378)	&	(294)	&	(7.63)	&	(5.09) \\
            
                                & \textit{SMAPE} &	45.08	&	41.05	&	41.71	&	41.31	&	46.57	&	43.68	&	41.89	&	42.14	&	41.23	&	158.1	&	66.09	&	55.03	&	42.74	&	\underline{\textbf{37.93}} \\
                                &&	(0)	&	(0)	&	(0)	&	(0)	&	(3.76)	&	(3.27)	&	(0.01)	&	(0.20)	&	(5.58)	&	(21.2)	&	(28.7)	&	(36.3)	&	(0.55)	&	(0.15) \\ \hline
        \multirow{5}{*}{Venezuela}   
                                & \textit{MASE} & 3.276	&	3.387	&	3.270	&	3.273	&	3.512	&	3.572	&	3.267	&	3.269	&	4.557	&	6.996	&	6.443	&	4.912	&	3.27	&	\underline{\textbf{3.096}} \\
                                && (0)	&	(0)	&	(0)	&	(0)	&	(0.07)	&	(0.07)	&	(0.001)	&	(0.001)	&	(1.19)	&	(0.98)	&	(1.49)	&	(0.96)	&	(0.01)	&	(0.14) \\
                                & \textit{RMSE} & 624.9	&	614.9	&	622.3	&	622.9	&	691.7 &	705.3	&	621.4	&	622.0	&	854.3	&	1240	&	1351	&	927.3	&	621.8	&	\underline{\textbf{592.7}} \\
                    Dengue            && 	(0)	&	(0)	&	(0)	&	(0)	&	(16.7)	&	(13.6)	&	(0.16)	&	(0.09)	&	(199)	&	(141)	&	(283)	&	(148)	&	(1.67)	&	(26.8) \\
        
                                & \textit{SMAPE} &  39.93 &	40.61	&	39.85	&	39.89	&	43.34	&	44.24	&	39.81	&	39.84	&	62.84	&	126.8	&	74.68	&	70.38	&	39.86	&	\underline{\textbf{37.01}} \\
                                && (0)	&	(0)	&	(0)	&	(0)	&	(0.98)	&	(1.01)	&	(0.01)	&	(0.01)	&	(21.7)	&	(33.7)	&	(24.6)	&	(23.9) &	(0.09)	&	(2.06) \\ \hline
       \multirow{5}{*}{Venezuela}   
                                & \textit{MASE} & 1.541	&	1.388	&	1.279	&	1.278	&	1.380	&	3.577	&	1.263	&	1.207	&	2.381	&	7.128	&	19.03	&	4.256	&	1.275	&	\underline{\textbf{1.067}} \\
                                && (0)	&	(0)	&	(0)	&	(0)	&	(0.45)	&	0.17	&	(0.001)	&	(0.03)	&	(1.04)	&	(0.89)	&	(47.9)	&	(1.10)	&	(0.02)	&	(0.004) \\
                                & \textit{RMSE} & 299.7	&	271.1	&	250.2	&	249.9	&	273.0	&	658.3	&	247.2	&	237.5	&	464.7	&	1208	&	4941	&	745.9	&	249.3	&	\underline{\textbf{218.4}} \\ 
                    Malaria             && 	(0)	&	(0)	&	(0)	&	(0)	&	(87.0)	&	(33.2)	&	(0.16)	&	(4.65)	&	(189)	&	(145)	&	(13$E^3$)	&	(177)	&	(3.26)	&	(0.72) \\
       
                                & \textit{SMAPE} &	18.25	&	16.30	&	14.95	&	14.94	&	16.48	&	50.00	&	14.76	&	14.08	&	24.05	&	138.5	&	57.77	&	64.75	&	14.90	&	\underline{\textbf{12.46}} \\
                                && (0)	&	(0)	&	(0)	&	(0)	&	(5.89)	&	(3.09)	&	(0.01)	&	(0.37)	&	(8.55)	&	(29.8)	&	(46.6)	&	(27.2)	&	(0.19)	&	(0.04) \\ \hline                                   
       \multirow{5}{*}{US-EPU}   
                                & \textit{MASE} & 1.595	&	1.685	&	1.671	&	1.948	&	1.681	&	1.617	&	1.658	&	1.655	&	1.837	&	2.511	&	2.758	&	2.421	&	1.591	&	\underline{\textbf{1.483}} \\
                                && 	(0)	&	(0)	&	(0)	&	(0)	&	(0.02)	&	(0.03)	&	(0.002)	&	(0.001)	&	(0.15)	&	(1.06)	&	(1.28)	&	(0.67)	&	(0.06)	&	(0.003) \\
                                & \textit{RMSE} & 81.41	&	85.73	&	85.03	&	93.35	&	85.53	&	82.19	&	84.50	&	84.34	&	89.77	&	110.2	&	124.9	&	108.9	&	80.93	&	\underline{\textbf{68.41}} \\
                    Index            && 	(0)	&	(0)	&	(0)	&	(0)	&	(0.70)	&	(1.99)	&	(0.05)	&	(0.02)	&	(4.64)	&	(34.5)	&	(57.5)	&	(20.9)	&	(2.75)	&	(0.15) \\
            
                                & \textit{SMAPE} & 29.24	&	31.36	&	31.01	&	38.27	&	31.27	&	29.77	&	30.72	&	30.64	&	35.53	&	59.35	&	59.99	&	53.67	&	29.19	&	\underline{\textbf{26.95}} \\
                                &&	(0)	&	(0)	&	(0)	&	(0)	&	(0.46)	&	(0.76)	&	(0.03)	&	(0.01)	&	(4.01)	&	(38.7)	&	(27.2)	&	(25.6)	&	(1.27)	&	(0.07)\\ \hline
    \multirow{5}{*}{UK unem-}   
                                & \textit{MASE} & 12.69	&	1.671	&	9.233	&	1.802	&	3.381	&	5.650	&	9.225	&	9.235	&	12.28	&	5.191	&	15.33
                                &	10.16	&	9.227	&	\underline{\textbf{1.465}} \\
                                && (0)	&	(0)	&	(0)	&	(0)	&	(1.17)	&	(0.02)	&	(0.01)	&	(0.002)	&	(4.08)	&	(2.89)	&	(5.21)	&	(2.23)	&	(0.05)	&	(0.13) \\
                                & \textit{RMSE} & 0.687	&	0.096	&	0.518	&	0.103	&	0.196	&	0.320	&	0.518	&	0.518	&	0.780	&	0.274	&	1.002	&	0.503	&	0.518	&	\underline{\textbf{0.082}} \\
                    ployment            && 	(0)	&	(0)	&	(0)	&	(0)	&	(0.07)	&	(0.002)	&	(0.001)	&	(0.001)	&	(0.28)	&	(0.15)	&	(0.33)	&	(0.10)	&	(0.002)	&	(0.005) \\
            
                                & \textit{SMAPE} &	11.31	&	1.564	&	9.497	&	1.691	&	3.175	&	5.259	&	9.490	&	9.499	&	10.73	&	4.822	&	13.80	&	9.100	&	9.492	&	\underline{\textbf{1.362}} \\
                                && 	(0)	&	(0)	&	(0)	&	(0)	&	(1.09)	&	(0.02)	&	(0.01)	&	(0.002)	&	(3.27)	&	(2.59)	&	(4.51)	&	(1.89)	&	(0.05)	&	(0.12) \\ \hline                           
        \multirow{5}{*}{Russia}   
                                & \textit{MASE} & 3.749	&	5.376	&	4.676	&	4.833	&	4.754	&	5.120	&	4.625	&	4.648	&	4.199	&	8.792	&	10.97	&	7.442	&	4.622	&	\underline{\textbf{3.162}} \\
                                && (0)	&	(0)	&	(0)	&	(0)	&	(0.20)	&	(0.05)	&	(0.003)	&	(0.001)	&	(0.83)	&	(5.96)	&	(11.9)	&	(5.60)	&	(0.11)	&	(0.69) \\
                                & \textit{RMSE} & 7.616	&	10.78	&	9.402	&	9.717	&	9.561	&	10.28	&	9.297	&	9.340	&	8.550	&	17.34	&	24.45	&	14.67	&	9.320	&	\underline{\textbf{6.555}} \\
            
                Exchange            && (0)	&	(0)	&	(0)	&	(0)	&	(0.39)	&	(0.10)	&	(0.01)	&	(0.001)	&	(1.58)	&	(10.8)	&	(27.3)	&	(10.4)	&	(0.17)	&	(1.32) \\
            
                                & \textit{SMAPE} & 10.04	&	14.75	&	12.69	&	13.15	&	12.92	&	13.99	&	12.55	&	12.61	&	11.36	&	27.37	&	31.26	&	22.78	&	12.55	&	\underline{\textbf{8.405}}\\
                                && 	(0)	&	(0)	&	(0)	&	(0)	&	(0.58)	&	(0.16)	&	(0.01)	&	(0.001)	&	(2.37)	&	(23.9)	&	(34.7)	&	(21.5)	&	(0.31)	&	(1.92) \\ \hline
         \multirow{6}{*}{Tourism}   
                                & \textit{MASE} & 0.726	&	0.801	&	\underline{\textbf{0.692}}	&	0.907	&	0.744	&	0.935	&	0.703	&	0.792	&	1.050	&	8.353	&	6.565	&	8.217	&	0.694	&	0.771\\
                                && 	(0)	&	(0)	&	(0)	&	(0)	&	(0.05)	&	(0.02)	&	(0.001)	&	(0.001)	&	(0.05)	&	(0.12)	&	(5.86)	&	(0.23)	&	(0.001)	&	(0.003)\\
                                & \textit{RMSE} & \underline{\textbf{72.53}}	&	87.48	&	75.21	&	95.90	&	83.25	&	99.46	&	78.11	&	80.28	&	94.55	&	639.1	&	587.4	&	628.9	&	75.99	&	85.66 \\
                                && (0)	&	(0)	&	(0)	&	(0)	&	(5.04)	&	(1.29)	&	(0.05)	&	(0.01)	&	(7.28)	&	(8.90)	&	(510)	&	(16.9)	&	(0.12)	&	(0.34)\\
                                & \textit{SMAPE} & 	8.327	&	9.262	&	\underline{\textbf{7.928}}	&	10.60	&	8.556	&	10.97	&	8.050	&  9.143	&	11.86	&	191.8	&	77.19	&	185.8	&	7.941	&	8.896\\
                                && (0)	&	(0)	&	(0)	&	(0)	&	(0.66)	&	(0.21)	&	(0.01)	&	(0.004)	&	(0.52)	&	(5.35)	&	(71.5)	&	(9.79)	&	(0.01)	&	(0.04) \\ \hline 
        
    \end{tabular}\label{L-finaltable}
\end{table*}
% \end{landscape}

\subsection{Benchmark Comparsions}

\begin{figure}
	\centering
	\includegraphics[scale=0.35]{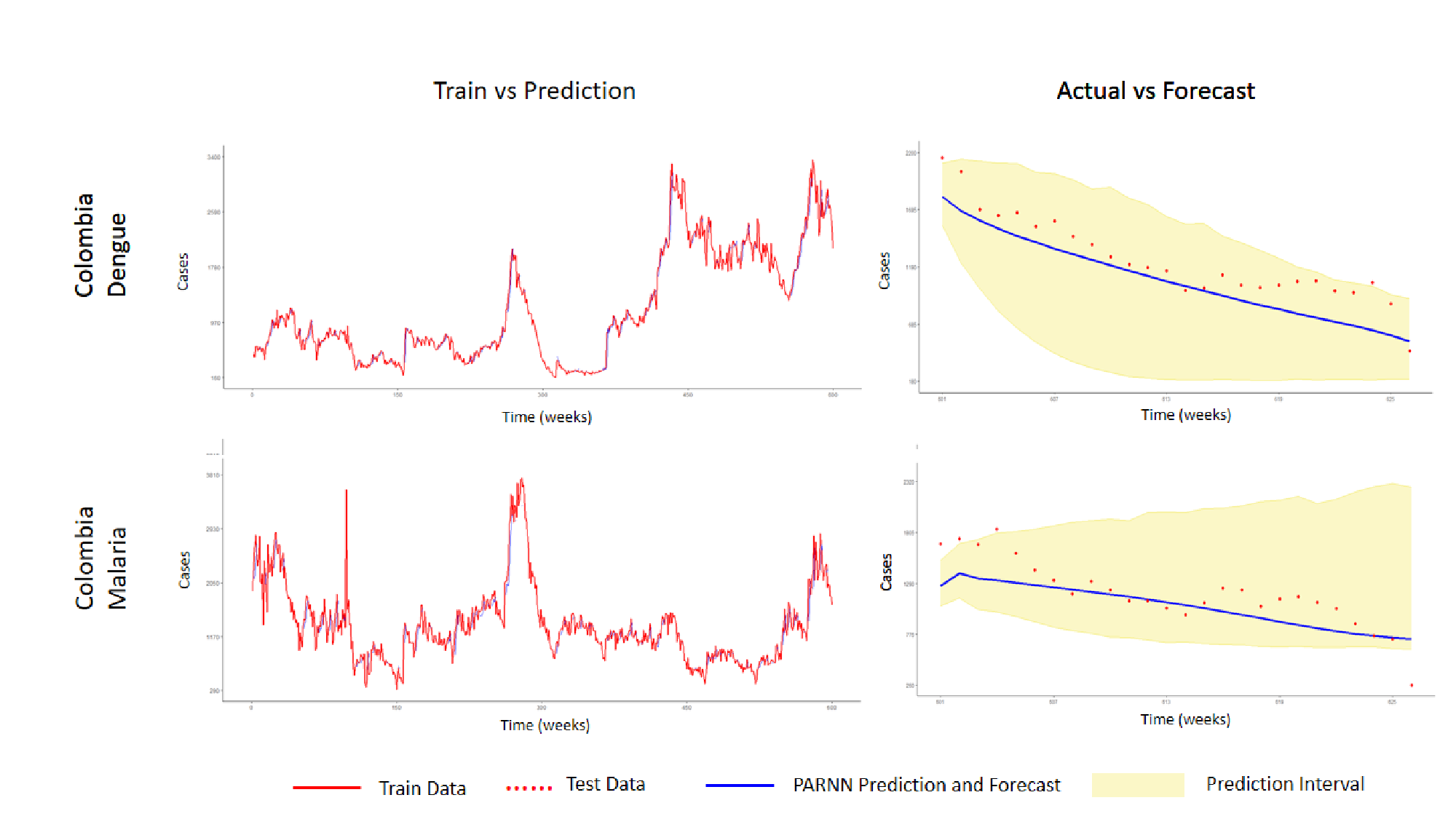}
	\caption{The plot shows input series (red line), ground truth (red points), 80\% prediction interval (shaded region), predictions (blue), and point forecasts (blue) generated by the PARNN model for the selected datasets.}
	\label{PARNN_CI}
\end{figure}

We evaluated the performance of the proposed PARNN model and thirteen benchmark forecasters for different forecast horizons. The experimental results for the short, medium, and long-term horizons are summarized in Tables \ref{S-finaltable}-\ref{L-finaltable}, respectively. It is observable from the given tables that there is an overall drop in the performance of all the baseline forecasters with the increasing forecast horizon since the more elongated the horizon more difficult the forecast problem. However, in the case of the proposed PARNN model, the long-range forecasting performance has significantly improved as evident from the experimental results. 
%Moreover, owing to the distinct scale and variance of the different datasets, there is a significant difference between their corresponding accuracy metrics. We can infer from the performance measures of different methods that the proposed PARNN model outperforms other baseline forecasters on most tasks. 
The proposal generates the best short-term forecast for daily datasets except for AMZN Stock data, where the MLP model demonstrates significant improvement compared to the other forecasters. However, for the medium-term forecasts of these datasets, the statistical forecasting technique, namely ETS, shows competitive performance alongside the proposal. For the long-term counterparts, the proposed PARNN framework can successfully diminish the MASE score by $20.05\%$, $60.34\%$, $25.96\%$, and $1.37\%$ for AMZN Stock, Births, MSFT Stock, and GOOG Stock prediction task, respectively in comparison with ARNN model. This improvement in the forecast accuracy of  the proposed model is primarily attributed to the augmentation of the input series with ARIMA residuals, which enables the model to generate accurate long-range forecasts. Furthermore, in the case of the epidemiological forecasting in the Venezuela region, we notice that statistical ETS and persistence-based RWD models outperform the benchmarks in the case of short-term forecasting, however, for the 26-weeks and 52-weeks ahead counterparts, the results generated by the proposed framework lie closer to the actual incidence cases. In the case of the dengue incidence in the Colombia region, the data-driven NBeats model forecasts the 13-weeks ahead disease dynamics more accurately as measured by the MASE and RMSE metrics, whereas the forecast generated by the conventional TBATS model is more accurate in terms of the relative error measure. On the contrary, for medium-term and long-term forecasting, the proposed PARNN framework outperforms the baseline models in terms of all the key performance indicators. A similar pattern is also prominent for the malaria cases of Colombia region, where the proposal generates the most reliable forecast for all the horizons. Furthermore, in the case of the macroeconomic datasets, namely US EPU Index series we observe that the deep neural architecture-based Transformers model generates competitive forecasts with the proposed framework for the medium-term forecasting analog. However, the baseline performance significantly deteriorates for the short-term and long-term horizons in comparison to the PARNN model. In the case of the Tourism dataset, the performance of the conventional RWD, ETS, and ARIMA models is significantly better than the other forecasters. However, the forecasts generated by the proposed PARNN model for the UK unemployment dataset outperform all the state-of-the-art forecasters as observed from the experimental evaluations. Moreover, for the Russia Exchange dataset, although the statistical TBATS model generates more accurate 6-months ahead forecasts, but for the 12-month and 24-month horizons the performance of the proposal is significantly better than the competitive models. In general, from the overall experimental results, we can infer that the proposed PARNN model can efficiently model and forecast complex time series datasets, especially for long-range predictions. Furthermore, the approximate run time of the proposed model is significantly less than other deep learning-based methods, and it also controls the model size to prevent the problem of overfitting. Finally, we display the medium-term forecast and the 80\% prediction interval generated by our proposed PARNN architecture for selected datasets in Fig. \ref{PARNN_CI}. From a visualization viewpoint, we can infer that the proposed PARNN overall show accurate and consistent forecasts compared to other state-of-the-art. 

\subsection{Significance Test}
In this section, we discuss the significance of improvements in accuracy metrics by performing different statistical tests. We initially conduct the non-parametric Friedman test to determine the robustness of different benchmarks and the proposed forecaster \cite{friedman1937use}. This statistical methodology tests the null hypothesis that the performance of different models is equivalent based on their average ranks for various datasets and rejects the hypothesis if the value of the test statistic is greater than the critical value. In this study, we assign rank $\tilde{r}_{i,j}$ to the $i^{th}$ model (out of $\kappa$ models) for its prediction task of the $j^{th}$ dataset (out of $\zeta$ datasets). The Friedman test computes the average rank of each forecaster as $\tilde{R}_i = \frac{1}{\zeta} \sum_{j=1}^{\zeta} \tilde{r}_{i,j}$. Under the null hypothesis, i.e., $\tilde{R}_i$ are equal for all $i = 1,2,\ldots,\kappa$ the modified version of Friedman test statistic given by:

\begin{equation*}
    F_{\mathcal{F}} = \frac{(\zeta-1)\chi^2_{\mathcal{F}}}{\zeta(\kappa-1)-\chi^2_{\mathcal{F}}} \;\; 
    \text{ where} \; \; \chi^2_{\mathcal{F}} = \frac{12\zeta}{\kappa(\kappa+1)}\left[ \sum_{i=1}^{\kappa} \tilde{R}_i^2 - \frac{\kappa(\kappa+1)^2}{4} \right],
\end{equation*}
follows a $F$ distribution with $(\kappa-1)$ and $(\kappa-1)(\zeta-1)$ degrees of freedom \cite{friedman1937use}. Following the Friedman test procedure, we compute the value of the test statistics for the 14 forecasting models across different test horizons of the 12 datasets and summarize them in Table \ref{Fried_test}.
% \begin{figure*}
% 	\centering
% 	\includegraphics[scale=0.55]{PARNN_Fit.png}
% 	\caption{The plot shows the input (black), ground truth (blue), and long-term forecast generated by PARNN (red), TBATS (green), and ARNN (yellow) models for selected datasets.}
% 	\label{actual vs predict}
% \end{figure*}

\begin{figure}
     \centering
     \begin{subfigure}{0.3\textwidth}
         \centering
         \includegraphics[scale=0.2]{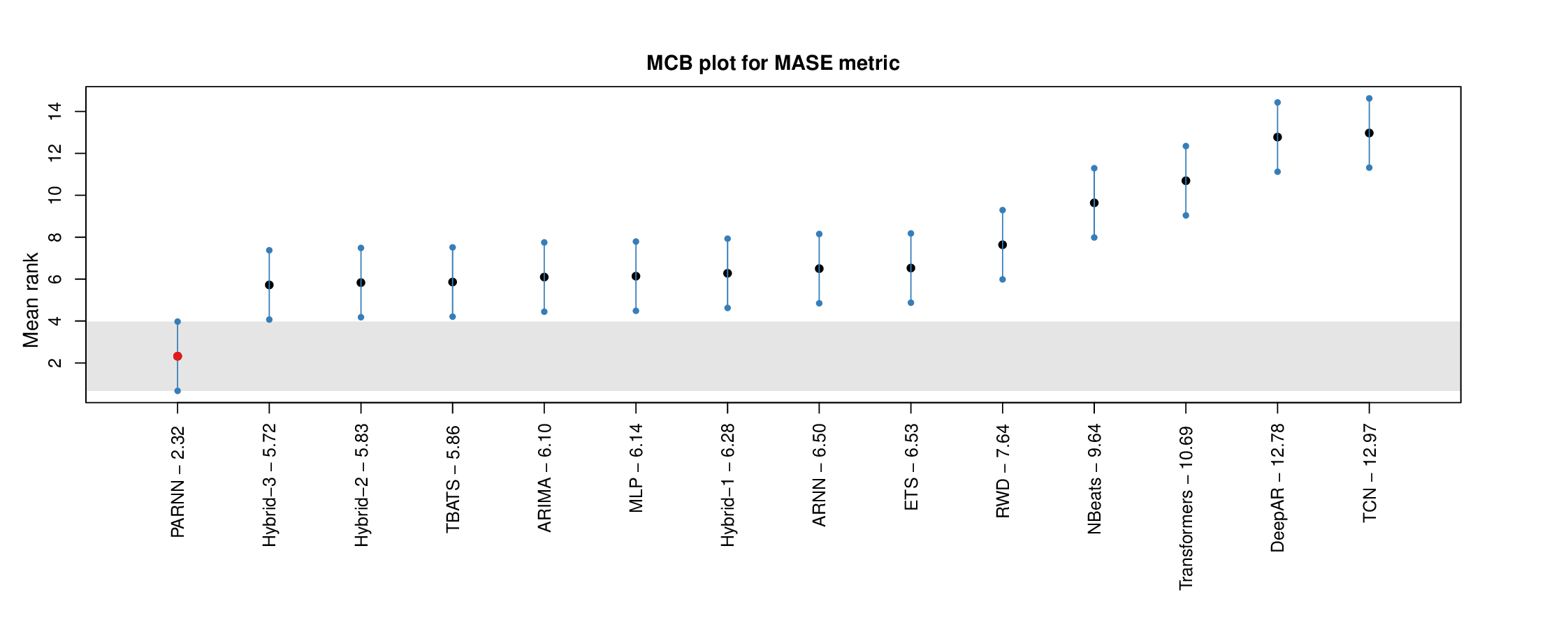}
         \caption{MASE MCB}
         %\label{fig:y equals x}
     \end{subfigure}
     \hfill
     \begin{subfigure}{0.3\textwidth}
         \centering
         \includegraphics[scale=0.2]{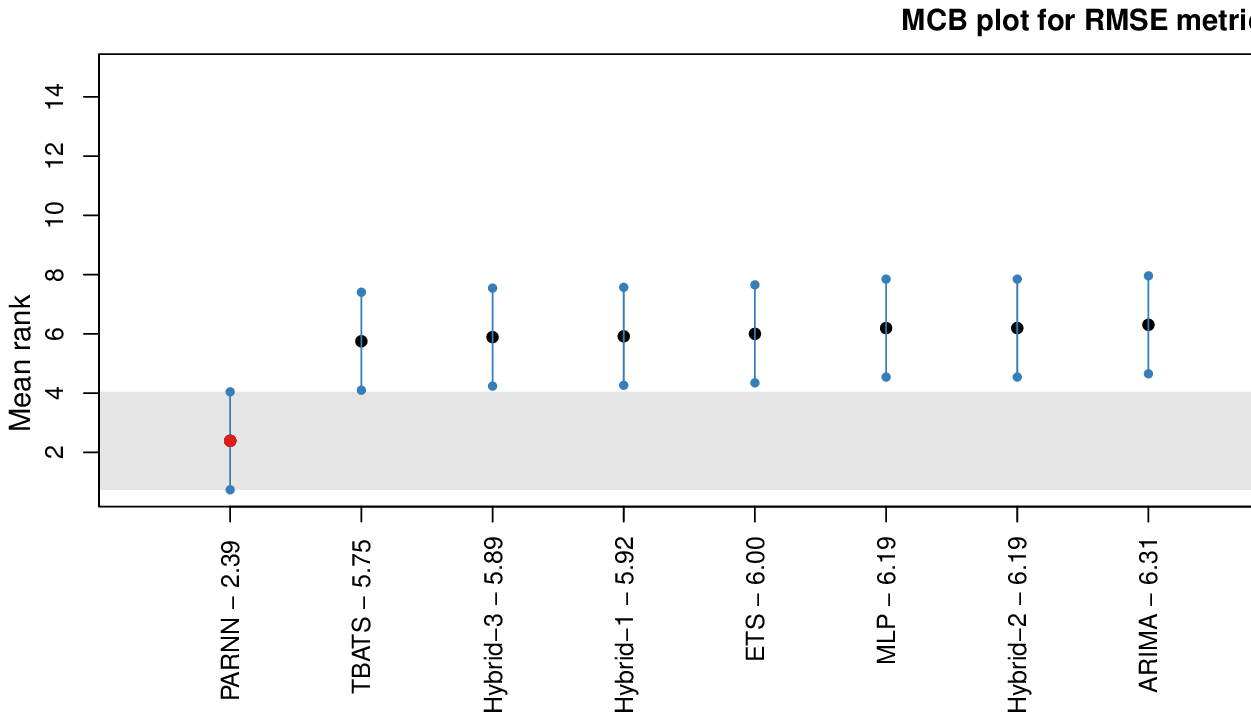}
         \caption{RMSE MCB}
     \end{subfigure}
        \caption{The plot shows the results of the MCB test. In this figure, PARNN-2.32 means that the average rank of the proposed PARNN model based on the MASE score is 2.32 when tested on all the datasets, similar for others.}
        \label{MCB_Results}
\end{figure}

\begin{table}[H]
    \centering
    \caption{Observed values of Friedman Test statistic for different accuracy metrics}
    \begin{tabular}{|c|c|c|c|} \hline
    Test Statistic & MASE & RMSE & SMAPE \\ \hline
    $\chi_{\mathcal{F}}^2$ & 237.2 & 236.2 & 250.9 \\ \hline
    $F_{\mathcal{F}}$  &  35.96 & 35.67 & 40.47 \\ \hline
    \end{tabular}
    \label{Fried_test}
\end{table}
Since the observed value of the test statistic $F_{\mathcal{F}}$ is greater than the critical value $F_{13,455} = 1.742$, so we reject the null hypothesis of model equivalence at a 5\% level of significance and conclude that the performance of the forecasters evaluated in this study is significantly different from each other. Furthermore, we utilize posthoc non-parametric multiple comparisons with the best (MCB) test \cite{koning2005m3} to determine the relative performance of different models compared to the `best' forecaster. To conduct this test,  we consider the MASE and RMSE metrics as the key performance indicator and compute the average rank and the critical distance for each model based on their respective scores. The results of the MCB test (presented in Fig. \ref{MCB_Results}) show that the PARNN model achieves the least rank, and hence it is the `best' performing model, followed by hybrid ARIMA-LSTM (Hybrid-3) and hybrid ARIMA-ARNN (Hybrid-2) models in case of the MASE metric. However, in terms of the RMSE scores, proposed PARNN framework has the minimum rank followed by the TBATS and hybrid ARIMA-LSTM (Hybrid-3) models. Moreover, the upper boundary of the critical interval of the PARNN model (marked by a shaded region) is the reference value for this test. As evident from Fig. \ref{MCB_Results}, the lower boundary of the critical interval for all other forecasters lies above the reference value, meaning that the performance of the baseline forecasters is significantly worse than that of the proposed PARNN model. Hence, based on the statistical tests, we can conclude that the performance of the proposed framework is considerably better in comparison to the baseline methodologies considered in this study.

\section{Conclusion and Discussion}
This paper presents a hybrid forecasting model that combines the linear ARIMA model with a nonlinear ARNN framework. This integration allows our proposed PARNN model to effectively handle various data irregularities such as nonlinearity, non-stationarity, and non-Gaussian time series. Importantly, PARNN operates within a "white-box-like" framework, eliminating the need for trial and error in choosing hidden units in the neural network. Through extensive experimentation on benchmark time series datasets from diverse domains, we have demonstrated that the PARNN model produces highly competitive forecasts across short, medium, and long-range horizons. Overall, we conclude that our proposed PARNN model introduces a valuable addition to the hybrid forecasting framework, without imposing implicit assumptions regarding the linear and nonlinear components of the dataset. This makes it highly suitable for handling real-world challenges faced by forecasting practitioners. Moreover, we presented a method for uncertainty quantification through confidence intervals, enhancing the model's richness. PARNN offers ease of use and efficiency compared to complex deep neural networks architectures like DeepAR, Transformer, NBeats, and TCN. Our experimental results consistently show the superiority of PARNN across various datasets with different frequencies and forecast horizons, especially for long-range forecasts. An immediate extension of this work would involve extending the PARNN model for multivariate time series forecasting and exploring the asymptotic behavior of the PARNN$(m, k, l)$ model.

\bibliographystyle{splncs04}
\bibliography{Bibliography}

\end{document}